%% file: arxiv.tex
\documentclass[10pt,twocolumn,letterpaper]{article}

\usepackage{iccv}
\usepackage{times}
\usepackage{epsfig}
\usepackage{graphicx}
\usepackage{amsmath}
\usepackage{amssymb}
\usepackage{subfigure}
\usepackage{color}
\usepackage{multirow}
\usepackage{textcomp, gensymb}
\usepackage{bbding}
\usepackage{booktabs}
\usepackage{bm, eucal}
\usepackage{pifont}
\usepackage{enumitem}

\usepackage{caption, multirow, overpic, textpos}
\usepackage{booktabs}
\usepackage[table]{xcolor}

\usepackage{xspace}
\usepackage{balance}
\usepackage{numprint}
\usepackage{sidecap}
\usepackage{float}
\usepackage{siunitx}
\usepackage{pifont}

\usepackage[pagebackref=true,breaklinks=true,letterpaper=true,colorlinks,bookmarks=false]{hyperref}

\usepackage[capitalize]{cleveref}
\crefname{section}{Sec.}{Secs.}
\Crefname{section}{Section}{Sections}
\Crefname{table}{Table}{Tables}
\crefname{table}{Tab.}{Tabs.}

\iccvfinalcopy 

\def\iccvPaperID{2663} 

\def\Ours{{Zolly}\xspace}
\def\Oursp{{$\rm Zolly^{\cP}$}\xspace}

\newcommand{\cP}{\mathcal{P}}

\begin{document}


\title{Zolly: Zoom Focal Length Correctly for Perspective-Distorted \\Human Mesh Reconstruction}

\author{Wenjia Wang$^{1, 4}$ \quad \; Yongtao Ge$^{2}$ \quad \; Haiyi Mei$^3$ \quad \; Zhongang Cai$^{3, 4}$ \quad \; \\ Qingping Sun$^3$ \quad \; Yanjun Wang$^3$ \quad \;  Chunhua Shen$^5$ \quad \; Lei Yang$^{3, 4, \dagger}$ \quad \; Taku Komura$^1$\\[1.5mm]
\normalsize $^1$ The University of Hong Kong \quad
\normalsize $^2$ The University of Adelaide \quad
\normalsize $^3$ SenseTime Research \\
\normalsize $^4$ Shanghai AI Laboratory \quad
\normalsize $^5$ Zhejiang University \quad\\
}
\ificcvfinal\thispagestyle{empty}\fi

\twocolumn[{%
	\renewcommand\twocolumn[1][]{#1}%
	\maketitle
	\begin{center}
		\newcommand{\teaserwidth}{\textwidth}
		\centerline{\includegraphics[width=\teaserwidth,clip]{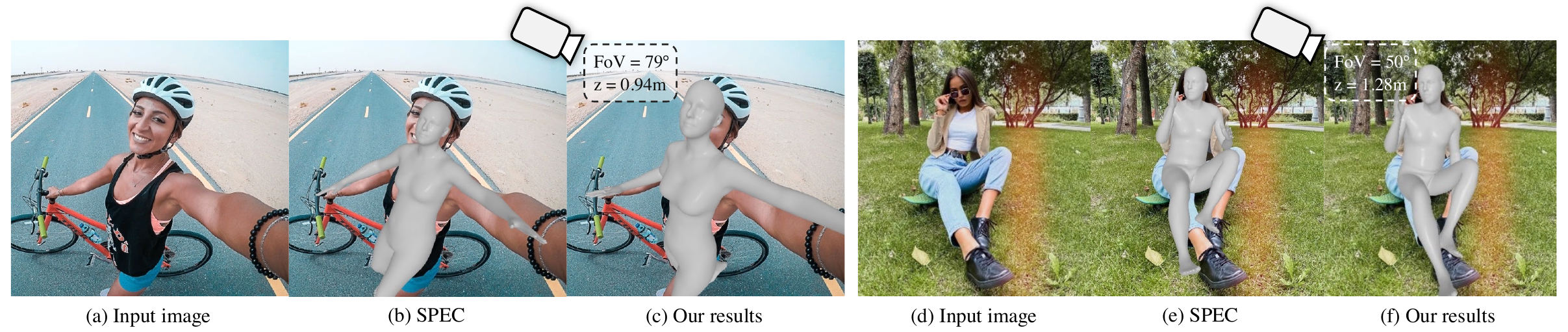}
		}
		\vspace{-1ex}
		\captionof{figure}{\label{fig:teaser}
 Close-up photography can make it difficult to discern 3D human pose in perspective-distorted images, while state-of-the-art methods often struggle with weak-perspective camera models or inaccurate focal length estimates. Our method overcomes these challenges, accurately recovering 3D human mesh from an approximate distance for fine-grained reconstruction. We could thus get focal length by our proposed approach.
  }
	\end{center}%
}]
{
  \renewcommand{\thefootnote}%
    {\fnsymbol{footnote}}
  \footnotetext[1]{LY$^\dagger$ is the corresponding author.}
}

\input{sections/0-abstract}

\input{sections/1-introduction}

\input{sections/2-related-work}

\input{sections/3-methods}

\input{sections/4-experiments_cr}

\input{sections/5-conclusion}



{\small
\bibliographystyle{ieee_fullname}
\bibliography{main}
}
\clearpage
\input{sections/6-appendix_cr}

\end{document}

%% file: sections/0-abstract.tex

\begin{abstract}
    As it is hard to calibrate single-view RGB images in the wild, existing 3D human mesh reconstruction~(3DHMR) methods either use a constant large focal length or estimate one based on the background environment context, which can not tackle the problem of the torso, limb, hand or face distortion caused by perspective camera projection when the camera is close to the human body. The naive focal length assumptions can harm this task with the incorrectly formulated projection matrices. To solve this, we propose \Ours, the first 3DHMR method focusing on perspective-distorted images. Our approach begins with analysing the reason for perspective distortion, which we find is mainly caused by the relative location of the human body to the camera center. We propose a new camera model and a novel 2D representation, termed distortion image, which describes the 2D dense distortion scale of the human body. We then estimate the distance from distortion scale features rather than environment context features. Afterwards, We integrate the distortion feature with image features to reconstruct the body mesh. To formulate the correct projection matrix and locate the human body position, we simultaneously use perspective and weak-perspective projection loss. Since existing datasets could not handle this task, we propose the first synthetic dataset PDHuman and extend two real-world datasets tailored for this task, all containing perspective-distorted human images. Extensive experiments show that \Ours outperforms existing state-of-the-art methods on both perspective-distorted datasets and the standard benchmark (3DPW). Code and dataset will be released at \url{https://wenjiawang0312.github.io/projects/zolly/}.

\end{abstract}

%% file: sections/1-introduction.tex
\section{Introduction}

\label{sec:intro}
Human pose and shape estimation from single-view RGB images is a long-standing research area of computer vision, as the reconstructed motion and mesh could empower various human-centered downstream applications like 3D animations, robotics, or AR/VR development. Previous works~\cite{hmr, spin, vibe, pare, romp, vibe, graphcmr, ochmr, ooh} formulate the problem under the assumption that the reconstructed people are far away from the camera, thus the torso and limb distortion caused by the perspective projection can be neglected.

However, perspective distortion in close-up images is common in real-life scenarios, such as photographs of athletes/actors in sports events/films or selfies taken for social media. In such images, distortions are usually caused by aerial photography, overhead beat, or large depth variance among torsos and limbs, resulting in depth ambiguity in single-view RGB images, which makes it a big challenge to recover human pose and shape (See \cref{fig:teaser}).

Previous methods typically assume a large fixed focal length~\cite{hmr,spin,vibe,pare} or estimate a focal length~\cite{spec} using pre-trained networks and calculate the translation from the estimated focal length. These settings are appropriate when people are far from the cameras, where the depth variance of the human body is negligible compared to the distance to the camera. However, these methods are inappropriate for handling scenarios in which human bodies are perspective distorted. Overestimating the focal length could lead to joint angle ambiguity or harm joint rotation learning.
Several methodologies for pose estimation, as proposed by previous works~\cite{beyondweak, cliff}, assume a large field-of-view (FoV) angle. However, these methods may not show significant improvement when the focus is solely on non-distorted human images, as they often lack a conditioning for depth variance when the camera zooms in or out. Inaccuracies in estimating the depth variance with respect to translation can adversely impact re-projection loss, leading to erroneous results, as illustrated in~\cref{fig:teaser}. Actually, a correctly estimated distance and focal length also help with 2D alignment, which will be useful in downstream tasks.

 To address the challenge of perspective distortion in close-up images, showing respect to Hitchcock's dolly zoom shot, we introduce \Ours~(\textbf{Z}oom f\textbf{O}cal  \textbf{L}ength correct\textbf{LY}) for perspective-distorted human mesh reconstruction. Our method utilizes 2D human distortion features to estimate the real-world distance to the camera center, enabling the reconstruction of the 3D human mesh in perspective-distorted images. The framework comprises of two parts: a translation estimation module for estimating the z-axis distance of the human body from the camera center, and a mesh reconstruction module for reconstructing 3D vertex coordinates in camera space. Additionally, we introduce a hybrid loss function that combines both perspective and weak perspective projection to boost performance.

\begin{figure}[t]
\begin{center}
   \includegraphics[width=0.85\linewidth]{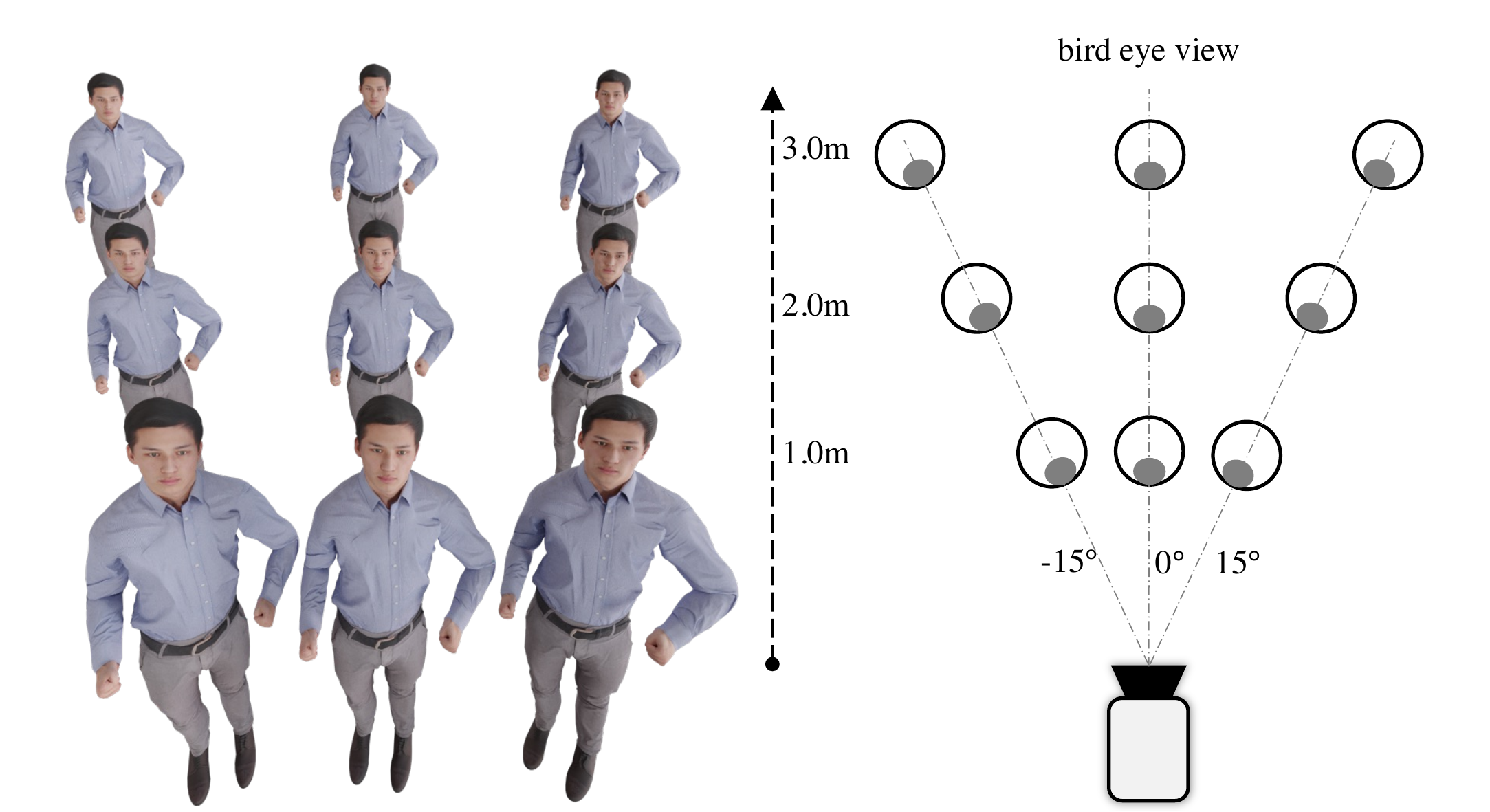}
\end{center}
   \caption{This figure showcases how the distortion scale of a person's limbs becomes more pronounced when closer to the camera center. Further, when two human bodies are at the same included angle with respect to the human-camera axis, they appear similar in facial direction. However, the horizontal translation may cause distinct distortion types on the left and right sides. We demonstrate that as the human body gets closer to the camera center, the distortion magnitude increases, leading to a more precise estimation of depth and rotation angles.}
\label{fig:dollyzoom}
\end{figure}

Inspired by the iconic dolly-zoom shot~\cite{jaws1975} (also known as zolly shot), which creatively combines camera movement and zooming to create a distorted perspective and sense of unease, we propose a translation estimation module for the perspective-distorted 3DHMR task. This module highlights how the relative position of the human body to the camera affects the perspective distortion in images. Based on this insight, we introduce the distorted image as a new representation to capture the 2D shrinking or dilation scales of each pixel. Our translation network utilizes distortion and IUV images to accurately estimate z-axis translation, overcoming the limitations of traditional methods that rely on environmental information. 
IUV image could help eliminate the 2D shift and scale information in distorted images and represent 2D dense position information.
For mesh reconstruction, we lift the 2D position feature to the 3D vertex position feature and sample the by-vertex distortion feature to regress 3D vertex coordinates. We use perspective projection to supervise correctly and weak-perspective projection  to locate the 2D human body position in the image and help to calculate our focal length.

In summary, our contributions are as follows:

(1) We analyze the state-of-the-art (SOTA) 3DHMR methods and propose a novel approach tailored to the perspective-distorted 3DHMR task.

(2) We propose a novel learning-based method to tackle the perspective-distorted 3DHMR task without relying on extra camera information. The core of our method is a newly designed representation, termed distortion image, and a hybrid projection supervision that make use of both perspective and weak-perspective projection.
(3) We build the first large-scale synthetic dataset PDHuman for the perspective-distorted 3DHMR task, with high-quality SMPL ground truth and camera parameters. To evaluate the performance on real images, we prepare two real-world benchmark datasets, SPEC-MTP~\cite{spec} and HuMMan~\cite{humman}, which contain perspective-distorted images with well-fitted SMPL parameters and camera parameters.


%% file: sections/2-related-work.tex
\section{Related Work}

\noindent\textbf{Mainstream 3D human mesh reconstruction methods.}
3D human pose estimation from a single RGB image is essentially an ill-posed problem. To obtain more realistic and manipulable human bodies, a parametric body model SMPL~\cite{smpl} was proposed, which uses 3D rotation representation to model human joint motions with defined LBS weights. To reconstruct human mesh from RGB images, there exist two mainstream pipelines: optimization-based methods and learning-based methods.

Optimization-based methods~\cite{smplify,fang2021mirrored} directly fit the body model parameters to 2D evidence via gradient back-propagation in an iterative manner.
Learning-based approaches~\cite{hmr,Pavlakos,decomr,pare} leverage a deep neural network to regress the human body model parameters or 3D coordinates of the human mesh, which can be further divided into model-based and model-free methods.
Inspired by 3D mesh reconstruction tasks\cite{c-axis, topdown, close, rfeps, global, avatarclip, sherf}, Model-based methods works~\cite{hmr,spec,pare,hybrik,smoothnet,deciwatch} utilize SMPL parameters to recover the human pose and shape. The milestone method HMR~\cite{hmr} takes it as a direct regression task.

Model-free works~\cite{graphcmr, meshgraphormer, fastmetro, tore} directly reconstructing 3D meshes from single view images. 

\begin{figure}[t]
\begin{center}
    \includegraphics[width=0.8\linewidth]{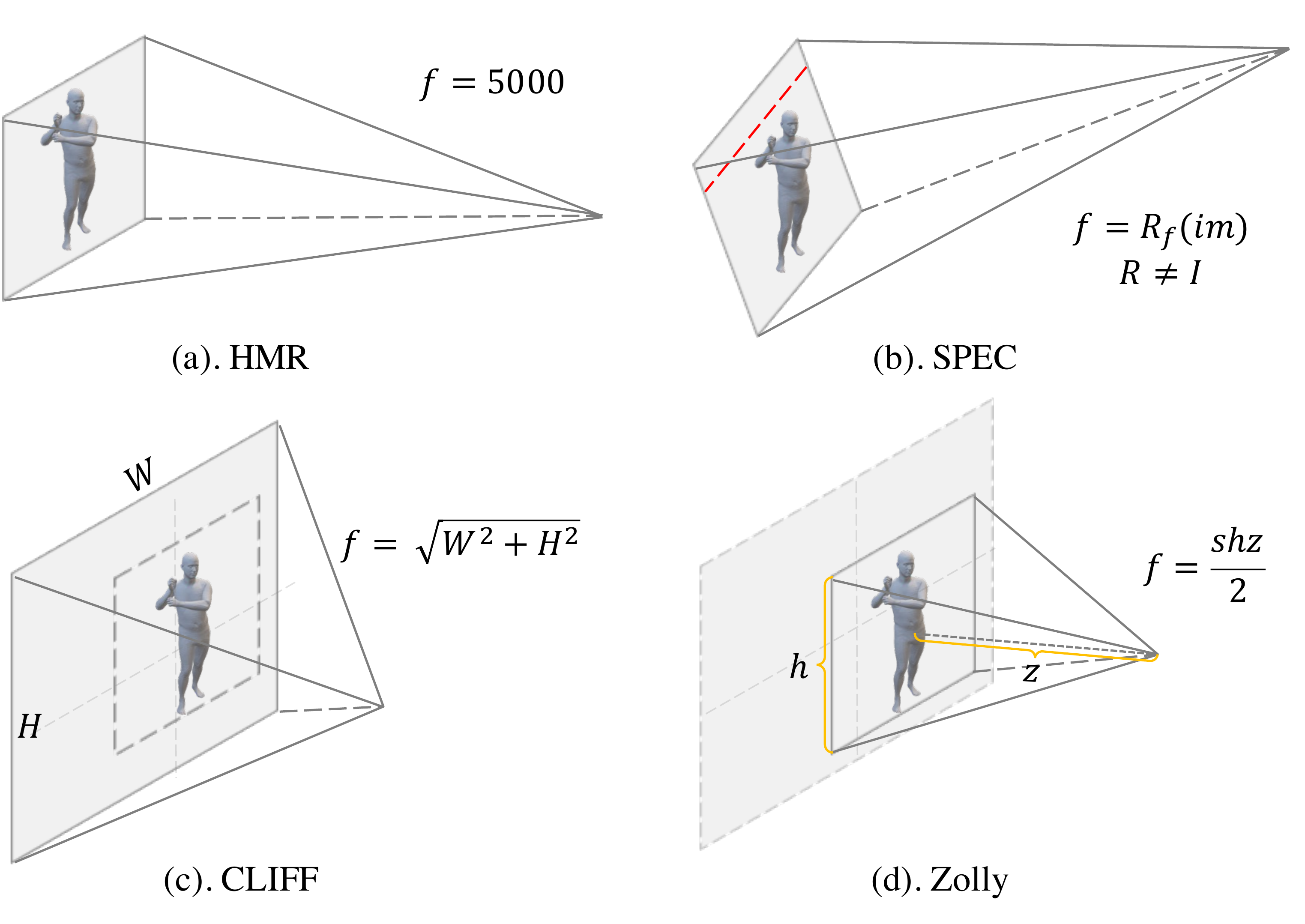}
\end{center}
\caption{
\textbf{Comparison of 4 different camera models in 3DHMR task}.
For HMR~\cite{hmr}, the focal length is fixed as 5000 pixels. Most methods follow this setting. For SPEC~\cite{spec}, the focal length is estimated by a network $R_{f}$ pre-trained on other datasets. For CLIFF~\cite{cliff}, during training, if without ground-truth focal length, will use the length of diagonal length. For \Ours, we use the estimated z-axis translation \emph{z}, camera parameter \emph{s}, and image height \emph{h} to calculate focal length.
}
\label{fig:camera}
\end{figure}

\noindent\textbf{Human mesh reconstruction with specific camera systems.} 
In the previous trend led by HMR~\cite{hmr}, the intrinsic camera model is formed as a weak-perspective camera, with a constant focal length of $5,000$ pixels. However, this assumption does not hold well when the person is close to the camera center, resulting in errors in the reconstructed 3D shape and pose. To address this issue, several recent works such as BeyondWeak~\cite{beyondweak}, SPEC~\cite{spec}, and CLIFF~\cite{cliff} have proposed different camera system assumptions. SPEC predicts camera parameters (pitch, yaw, and FoV) from a single-view image, but its asymmetric Softargmax-$\mathcal{L}_2$ loss tends to overestimate focal length and translation, which is not suitable for distorted images. Moreover, SPEC regresses camera parameters through environmental information, which can sometimes be meaningless when the background lacks geometry information. CLIFF focuses on joint rotation variance caused by horizontal shift but has not conditioned the distance from the human body to the camera. CLIFF, following BeyondWeak\cite{beyondweak}, uses the diagonal length of the image as the focal length, which is not a close assumption for distorted problems since the focal length can be easily adjusted during image capture.

Compared to these methods, our framework estimates the z-axis translation from 2D human distortion features, and obtains a more accurate focal length from the estimated translation, leading to much better reconstruction accuracy on distorted images. See a comparison of the camera models in \cref{fig:camera}.
In the Sup. Mat., we quantitatively demonstrate the bad re-projection influence caused by a wrongly formulated projection matrix.

%% file: sections/3-methods.tex
\section{Methodology}
\label{chapter:method}
In this section, we first review the formulation of previous camera systems and then present our camera system customized for distorted images in~\cref{chapter:method-camera}. 
~\cref{chapter:net_structure} presents our network architecture with two key components: (i) translation estimation module and (ii) mesh reconstruction module.
Subsequently, we explain the proposed hybrid re-projection loss functions for distorted human mesh reconstruction in~\cref{chapter:method-loss}.


    



\begin{figure*}[t]
  \centering
   \includegraphics[width=0.95\linewidth]{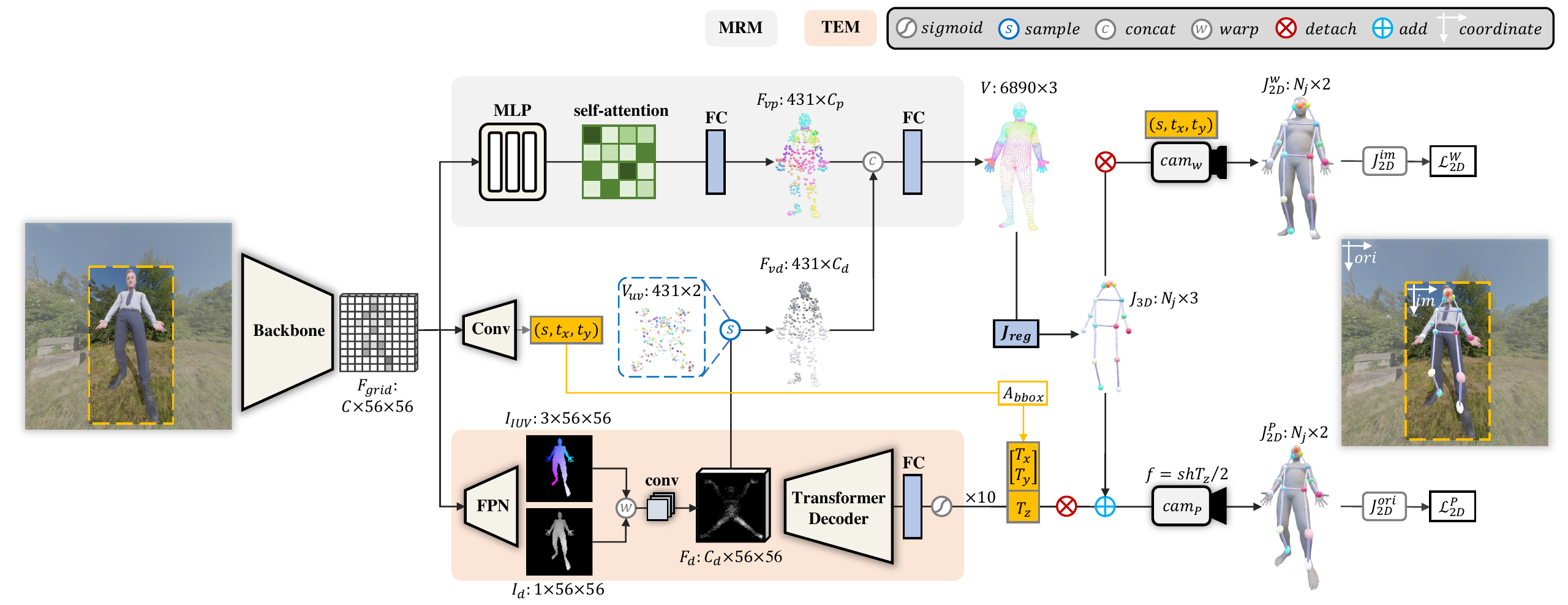}
  \hfill
  \caption{\textbf{\Ours pipeline overview}. The whole pipeline mainly consists of two modules and a hybrid re-projection supervision. MRM denotes the mesh reconstruction module. TEM indicates the translation estimation module. $F_{grid}$ is the spatial feature from the backbone. $F_{vp}$ and $F_{vd}$ represents per-vertex position and distortion feature . $(s, t_{x}, t_{y})$ are the weak-perspective parameters. $J_{2D}^{im}$ denotes 2D joints in the cropped image coordinate system. $J_{2D}^{ori}$ denotes 2D joints in original image coordinate system before cropped. $h$ denotes image height.}
  \label{fig:pipeline}
\end{figure*}


\subsection{Preliminary}
\label{chapter:method-camera}
\noindent\textbf{Camera system analysis.} In weak-perspective projection, the inner depth variance is ignored in the human body, which means this projection model views the human body as a planar object without thickness. Thus the projection matrix should be as follows:
\begin{equation}
\small
\begin{bmatrix}
  f&  & \\
  &f  & \\
  &  &1
\end{bmatrix}\begin{bmatrix}
 x+T_{x}\\
 y+T_{y}\\
z+T_{z}
\end{bmatrix} = \begin{bmatrix}
 f(x+T_{x})\\
 f(y+T_{y})\\
T_{z},
\end{bmatrix}\text{, $z=0$},
\end{equation}
where $f$ refers to the focal length in NDC (Normalized Device Coordinate) space, $x, y, z$ refers to a vertex point on human body mesh and $T_{x}, T_{y}, T_{z}$ refers to pelvis translation.
The weak-perspective camera parameters $(s, t_{x}, t_{y})$, which represent 2D orthographic transformation, could be used to approximate the projection:
\begin{equation}
\small
\begin{bmatrix}
f(x+T_{x})/T_{z} \\
f(y+T_{y})/T_{z}
\end{bmatrix}
= \begin{bmatrix}
s(x+t_{x})  \\
s(y+t_{y}) 
\end{bmatrix}.\\
\end{equation}
Finally, we can get:
\begin{equation}
\small
s \times T_{z} = f\text{, }T_{x} = t_{x}\text{, }T_{y} = t_{y}.
\label{equation:f_sz}
\end{equation}
However, the perspective projection actually is:
\begin{equation}
\small
\begin{bmatrix}
x_{2D} \\
y_{2D}
\end{bmatrix} = \begin{bmatrix}
f(x+T_{x})/(z+T_{z}) \\
f(y+T_{y})/(z+T_{z})
\end{bmatrix}
= \begin{bmatrix}
s(x+t_{x})  \\
s(y+t_{y}) 
\end{bmatrix}.
\label{equation:pers_proj}
\end{equation}
From~\cref{equation:pers_proj}, if $z$ gets smaller, the projected $x_{2D}, y_{2D}$ will be larger. This phenomenon causes the closer points on the close-up photographed image to dilate and the farther points to shrink. \textit{Thus, the 3D translation results in pixel-level distortion on the limbs, torso, or faces in the 2D projected image.} Usually, when the human body is farther than $5~\si{\metre}$, the distortion is subtle. Under these circumstances, a weak-perspective projection could be used.

Following the weak-perspective assumption, we take the $f=s \times T_{z}$ ($f$ is in NDC space) as an approximation. $T_{z}$ is the z-axis translation of the pelvis, which could be viewed as a mean translation of the whole human body. The difference compared to previous methods~\cite{hmr, spec, cliff}
is that we first estimate the body translation and then calculate the focal length.    
So we still need to estimate weak-perspective camera parameters $(s, t_{x}, t_{y})$ to compute the focal length and obtain the 2D location in the image. Following SPEC~\cite{spec}, we get the $T_{x}, T_{y}$ in the full image from the $t_{x}, t_{y}$ by affine transformation using the bounding box. 

\noindent\textbf{Distortion image.}
As described in~\cref{chapter:method-camera}, our approach projects the 3D translations of human body points into 2D images where the limbs dilate or shrink. We adopt the $x$-$y$ plane of the pelvis as the reference plane representing a `scale equals 1' plane. When the human is distant from the camera, it can be approximated as a zero-thickness plane, where all distortion scales are 1. And as shown in~\cref{equation:pers_proj}, distortion scales are inversely proportional to the z-distance from the camera when the body is closer. To quantify limb distortion caused by perspective projection, we introduce a distortion image $I_d$, where $I_d= T_z / I_{Depth}$, where $I_{Depth}$ represents a depth image. The distorted image and its pixel value enable a visual and numerical representation of the limb dilation or shrinkage caused by the perspective camera. For instance, when the pelvis is fixed, a finger can appear twice as dilated when $z$ of the finger reaches from \SIrange{1}{0.5}{\meter}. See \cref{fig:dataset} for the demonstration of distortion image.
%

\subsection{Network Structure}
\label{chapter:net_structure}

Given a monocular image, \Ours applies an off-the-shelf Convolution Neural Network, \eg~\cite{resnet,hrnet}, as an image encoder; the output multi-level features can be used as an input for the translation module and the mesh reconstruction module that we describe next.

\noindent\textbf{Translation module.} As shown in~\cref{fig:pipeline}, we estimate the distortion image $I_{d}$ and IUV image $I_{IUV}$ with an FPN~\cite{FPN} structure. 
There are two main advantages for this setting. Firstly, we can distillate the geometry information on the human body without background context.
Another advantage is that this dense correspondence predicting task is easy to train.
As noted in~\cref{chapter:method-camera}, the distortion type corresponds to one certain translation, and the distortion is determined when the image was captured, whether or not cropped or rotated afterwards. So we further warp the distorted image into the continuous UV space~\cite{decomr} to eliminate the 2D scale, shift, and rotation. 
We treat $T_{z}$ as a learnable embedding.
A $1\times1$ convolution is first applied to up-sample the channels of the warped distortion image. Then cross-attention~\cite{transformer} is performed between the warped distortion feature and the z-axis embedding, with a fully connected layer to output $T_{z}$. Note that we use sigmoid then $\times10$ to restrict $T_{z}$ to be between 0\si{\metre} and 10\si{\metre}.
Following SPEC~\cite{spec}, we get $T_{x}$ and $T_{y}$ by applying affine transform on the estimated $t_{x}, t_{y}$ with ground-truth bounding boxes (See Sup. Mat. for more details).
The loss function of the translation module is formulated as follows:
\begin{equation}
\small
\mathcal{L}_{Transl} = \lambda_{IUV}\mathcal{L}^{2}_{IUV} + \lambda_{d}\mathcal{L}^{2}_{d}  + \lambda_{z}\mathcal{L}^{1}_{z},
\end{equation}
where $\mathcal{L}_{IUV}^{2}$ is the $\mathcal{L}2$ loss of the IUV image, $\mathcal{L}_{d}^{2}$ is the $\mathcal{L}2$ loss of the distortion image, and $\mathcal{L}_{z}^{1}$ is the $\mathcal{L}1$ loss of z-axis translation.

\noindent\textbf{Mesh reconstruction module.} 
Different from previous methods that use graph convolution~\cite{graphcmr} or transformers~\cite{fastmetro, meshgraphormer} for building long-range dependence among different vertices, we adopt a light-weight MLP-Mixer~\cite{mlpmixer} structure to model the attention among different vertices, followed by a fully connected layer that lifts per-vertex position features $F_{vp}$ from the spatial feature~$F_{grid}$ which was used to predict $I_{IUV}$. 

As illustrated in~\cref{fig:pipeline}, since the distortion feature has already been warped into UV space, we could easily sample the per-vertex distortion feature $F_{vd}$ from the warped distortion feature $F_d$ by pre-defined Vertex UV coordinates $V_{uv}$~\cite{decomr}.
We concatenate $F_{vd}$ with $F_{vp}$ and use fully connected layers to predict the coordinates of a coarse mesh of the body that is composed of $431$ vertices.
The coarse mesh is up-sampled using two fully connected layers, resulting in an intermediate mesh with $1,723$ vertices and a full mesh with $6,890$ vertices. 3D joint coordinates are obtained using a joint regression matrix provided by the SMPL~\cite{smpl} body model.
The total loss for the mesh reconstruction module is:
\begin{equation}
\small
\begin{split}
\mathcal{L}_{Mesh} = \lambda_{J_{3D}}\mathcal{L}^{1}_{J_{3D}} + \lambda_{J_{2D}^{P}}\mathcal{L}^{1}_{J_{2D}^{P}} + \lambda_{J_{2D}^{W}}\mathcal{L}^{1}_{J_{2D}^{W}}\\ + \lambda_{V}(\mathcal{L}^{1}_{V^{''}}+\mathcal{L}^{1}_{V^{'}}+\mathcal{L}^{1}_{V}) ,
\end{split}
\end{equation}
where $\mathcal{L}^{1}_{J_{3D}}$ is $\mathcal{L}$1 loss of 3D joints, $\mathcal{L}^{1}_{V^{''}}$, $\mathcal{L}^{1}_{V^{'}}$ and $\mathcal{L}^{1}_{V}$ is $\mathcal{L}$1 loss of coarse, intermediate vertices, and full vertices respectively. $\mathcal{L}^{1}_{J_{2D}^{P}}$ and $\mathcal{L}^{1}_{J_{2D}^{W}}$ represents loss of perspective and weak-perspective re-projected 2D joints, and will be further illustrated in~\cref{chapter:method-loss}.

\subsection{Hybrid Re-projection Supervision}


Most existing methods~\cite{hmr, cliff, fastmetro} usually use a pre-defined focal length $f$. SPEC~\cite{spec} train a CamCalib network to estimate the focal length.
Then, z-axis translation $T_{z}$ can be calculated by $T_{z} = 2f/hs$ .
On the contrary, as illustrated in~\cref{equation:f_sz}, we aim to get the focal length $f$ by directly predicting the orthographic scale $s$ and z-axis translation $T_{z}$. Following HMR~\cite{hmr}, we still use the weak-perspective projection besides perspective projection. 

\noindent\textbf{Weak-perspective re-projection.}
For weak-perspective projection loss, we follow HMR~\cite{hmr}, use focal length $ f_{W}$ as $5,000$ pixels, and thus formulate the weak-perspective intrinsic matrix and translation separately as:

\begin{equation}
\small
 \mathit{K}_{W} =\begin{bmatrix}
 f_{W} &  & h/2\\
  &   f_{W} & h/2\\
  &  & 1
\end{bmatrix}\text{, }
\mathit{T_{W}} = \begin{bmatrix}
   t_{x}  \\
  t_{y} \\
 2f_{W}/sh
\end{bmatrix}.
\end{equation}
Then we project the 3D joints $\hat{J}_\mathit{3D}$ and measure the difference with 2D keypoints in image coordinates as: 

\begin{gather}
\small
    \hat{J}_\mathit{2D}^{W} = K_{W}(\hat{J}_\mathit{3D}^{\otimes} + T_{W}),
\\
\mathcal{L}_{\mathit{2D}}^{W} = \sum_{i=1}^{N_{j}} \frac{1}{d_{J[i]}}\| \hat{J}_\mathit{2D}^{W}[i] \; - \; J_\mathit{2D}^{im}[i] \|_F^1,
\label{eq:weak_loss}
\end{gather}
where \textit{{ $\hat{J}_\mathit{3D}^{\otimes}$} means we detach the gradient from the body model joints in weak-perspective projection.} This means we only update the weak-perspective camera $(s, t_{x}, t_{y})$ and do not want this wrong projection to harm the body pose gradient flow. $(s, t_{x}, t_{y})$ are mainly used to locate the human body's position in image coordinates and compute the focal length ${f_{P}}$. For better position alignment, we divide a distortion weight $d_{J[i]}$, which is sampled from distortion image $I_{d}$ by $J^{im}_{2D}[i]$ for every joint. This forces the dilated limbs to get a smaller weight while the shrunk limbs get a bigger weight.

\paragraph{Perspective re-projection.}
\label{chapter:method-loss}
Perspective re-projection is mainly used to supervise pose or mesh reconstruction with the correct projection matrix.
Firstly, we have 3D joints by $\hat{J}_\mathit{3D} = \mathcal{J}_{reg}V$.
We use ground-truth focal length $ f_{P}$ to stabilize the training. For samples without ground-truth focal length, we will use a focal length of $1,000$ pixels for 224$\times$224 images. This will make the translation range approximately from 5 to 10 meters. During inference, according to ~\cref{equation:pers_proj}, we compute the focal length in screen space for perspective projection by $ f_{P} = shT_{z}/2$ pixels, where $h$ represents cropped image height, equals 224 pixels in our setting. Thus we can formulate the perspective intrinsic matrix $\mathit{K}_{P}$ and projected 2D joints $\hat{J}_\mathit{2D}^{P}$ as:
\begin{gather}
\small
\mathit{K}_{P} =\begin{bmatrix}
 f_{P} &  & H/2\\
  &   f_{P}  & H/2\\
  &  & 1
\end{bmatrix}, 
\hat{J}_\mathit{2D}^{P} = K_{P}(\hat{J}_\mathit{3D} + T_{P}^{\otimes}),
\label{eq:project}
\end{gather}
where $\mathit{T_{P}^{\otimes}}$ is the translation estimated by translation head in~\cref{chapter:net_structure}. We detach it as well to avoid the alignment conflicting of two re-projection.
We project the 3D joints $\hat{J}_\mathit{3D}$ and measure the difference with the original 2D keypoints in the image coordinates before cropped.

%% file: sections/4-experiments_cr.tex
\section{Experiments}

\subsection{Datasets}

\begin{figure}[t]
    \centering
  \includegraphics[width=0.93\linewidth]{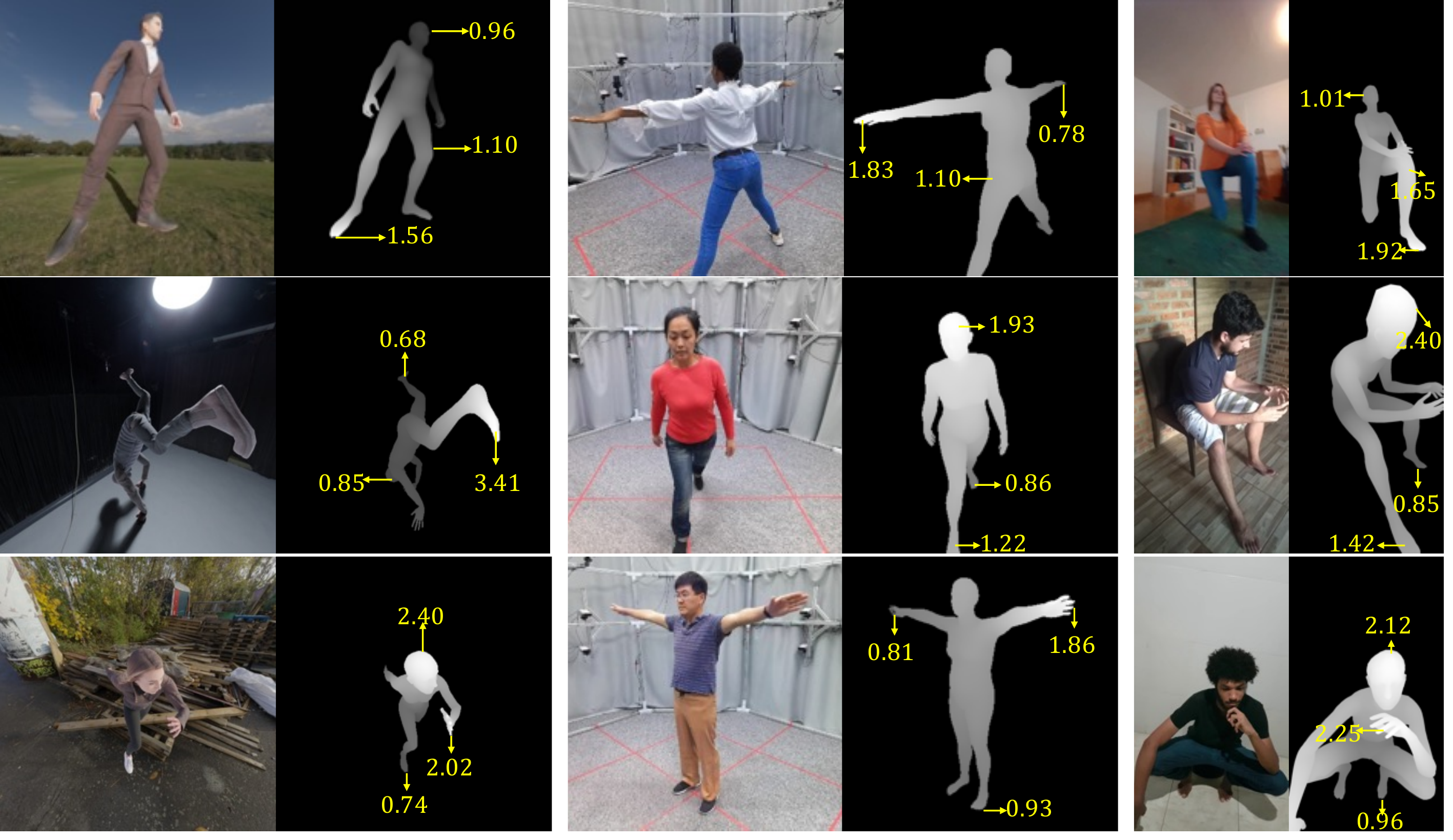}
    \caption{\textbf{Demonstration of distortion image.} The three columns from left to right are our PDHuman dataset, HuMMan dataset, and SPEC-MTP dataset, respectively. The value from the arrow (yellow) indicates the distortion scale of the pixel.}
    \label{fig:dataset}
\end{figure}

\noindent\textbf{PDHuman.}
Despite perspective distortion being a common problem, no existing public dataset is specifically designed for this task.
Inspired by recent synthetic datasets~\cite{synbody,gta,bedlam,Patel:CVPR:2021:AGORA}, we introduce a synthetic dataset named PDHuman. The dataset contains $126,198$ images in the training split and $27,448$ images in the testing split, with annotations including camera intrinsic matrix, 2D/3D keypoints, SMPL parameters ($\boldsymbol{\theta}$, $\boldsymbol{\beta}$), and translation for each image. The testing split is further divided into 5 protocols by the max distortion scale of each image sample.
We define the max distortion scale for each sample as $\tau$; this value will be used in splitting protocols.

We use $630$ human models from RenderPeople~\cite{renderpeople} and $1,710$ body pose sequences from Mixamo~\cite{mixamo}, with 500 HDRi images with various lighting conditions as backgrounds. 
We use the dolly-zoom effect to generate random camera extrinsic and intrinsic matrices with random rotations, translations, and focal lengths.
The distance from the human body to the cameras is set from 0.5m to 10m, so our dataset contains severely distorted, slightly distorted, and nearly non-distorted images.
Then we use Blender~\cite{blender} to render the RGB images. See \cref{fig:dataset} for brief demonstration.
For detailed rendering procedures and more image demonstrations, please refer to Sup. Mat.



\noindent\textbf{SPEC-MTP and HuMMan Datasets.}
SPEC-MTP dataset~\cite{spec} is proposed to test human pose reconstruction in world coordinates. It includes many close-up shots, mostly taken from below or overhead views, leading to images with distorted human bodies. HuMMan dataset~\cite{humman} is captured by multi-view RGBD cameras and has accurate ground truth because the SMPL parameters are fitted based on 3D keypoints and point clouds. HuMMan also contains images with distorted human bodies since the actors were close to the cameras, all less than 3 meters away. In our paper, we extend both datasets into real-world perspective-distorted datasets. SPEC-MTP is used only for testing. For HuMMan, we split it into training and testing parts. When testing, we divide these two datasets into three protocols based on their maximum distortion scale $\tau$. See \cref{fig:dataset} for brief demonstration.


\noindent\textbf{Non-distorted Datasets.}
For non-distorted datasets, we use Human3.6M~\cite{h36m}, COCO~\cite{coco}, MPI-INF-3DHP ~\cite{mpi-inf3dhp} and LSPET~\cite{lspet} as our training data.
Following~\cite{meshgraphormer, fastmetro}, we also report the results fine-tuned on 3DPW~\cite{3dpw} training data.

\subsection{Evaluation Metrics}
To measure the accuracy of reconstructed human mesh, we follow the previous works~\cite{hmr,spec,cliff} by adopting MPJPE (Mean Per Joint Position Error), PA-MPJPE (Procrustes Analysis Mean Per Joint Position Error) and PVE (Per Vertex Error) as our 3D evaluation metrics. They all measure the Euclidean distances of 3D points or vertices between the predictions and ground truth in millimeters (mm).

To measure the re-projection results in perspective distorted datasets such as PDHuman, SPEC-MTP and HuMMan, we leverage metrics widely used in segmentation tasks, MeanIoU~\cite{pascal} as our 2D metric. We both report foreground and background MeanIoU marked as mIoU and body part MeanIoU marked as P-mIoU. We use the 24-part vertex split provided by official SMPL~\cite{smpl} for body part segmentation. During the evaluation, for weak-perspective methods like HMR, we will render the predicted segmentation masks with a focal length of $5,000$ pixels. And we use the corresponding focal length on methods with specific camera models, such as SPEC~\cite{spec}, CLIFF~\cite{cliff}, and proposed \Ours.

\begin{table*}[t]
    \centering
    \scalebox{0.61}{\input{tables/sota-distortion}}
    
    \caption{Results of SOTA methods on PDHuman, SPEC-MTP~\cite{spec} and HuMMan~\cite{humman} datasets. Here we report the largest distortion protocol. 
    R50 terms ResNet-50~\cite{resnet}, and H48 terms HRNet-w48~\cite{hrnet} here.}
    \label{tab:sota}
\end{table*}

\subsection{Implementation Details}
Unless specified, we use ResNet-50~\cite{resnet} and HRNet-w48~\cite{hrnet} backbones for model-free \Ours. 
We also design a model-based variant, \Oursp ($\cP$ stands for parametric), by changing the mesh reconstruction module to a model-based pose and shape estimation module. The details of \Oursp can be found in the Sup. Mat.
%
All backbones are initialized by COCO~\cite{coco} key-point dataset pre-trained models.
We use Adam~\cite{adam} optimizer with a fixed learning rate of $2e^{-4}$. 
All experiments of \Ours are conducted on 8 A100 GPUs for around 160 epochs, 14$\sim$18 hours. 
 Our training pipeline was built based on MMHuman3D~\cite{mmhuman3d} code base.
For samples with ground-truth focal length and translations, we render IUV and distortion images online during the training by PyTorch3D~\cite{pytorch3d}.

For comparison on PDHuman, SPEC-MTP, and HuMMan, we follow the official codes of HMR~\cite{hmr}, SPEC~\cite{spec}, PARE~\cite{pare}, GraphCMR\cite{graphcmr}, FastMETRO\cite{fastmetro}. We re-implemented CLIFF~\cite{cliff} since the authors have not released the training codes. All SOTA methods are trained on 8 A100 GPUs until convergence, following the officially released hyper-parameters. 
All methods are trained on the same datasets with the same proportion, \eg, 
Human3.6M~\cite{h36m} (40\%), PDHuman (20\%), HuMMan~\cite{humman} (10\%), MPI-INF-3DHP~\cite{mpi-inf3dhp} (10\%), COCO~\cite{coco} (10\%), LSPET~\cite{lspet} (5\%).

For distorted datasets, we only report the results on the protocols with the largest distortion scales. The full results of all protocols are shown in Sup. Mat.

\subsection{Main Results}
\begin{figure*}[htp]
  \centering
\includegraphics[width=0.97\linewidth]{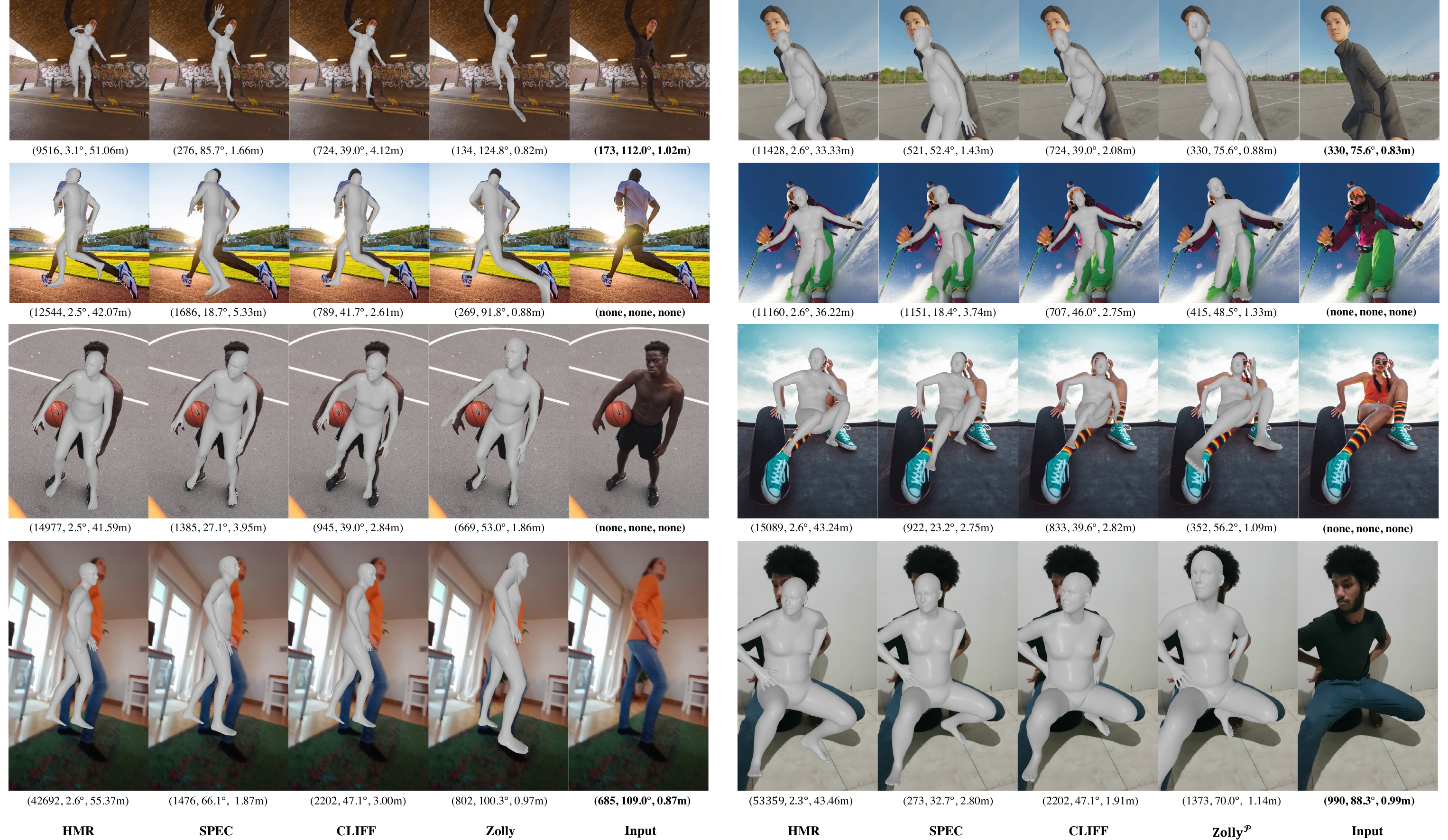}
  \caption{\textbf{Qualitative results of SOTA methods.} Besides \Ours, we visualize the results of three methods with specific camera models: HMR~\cite{hmr}, SPEC~\cite{spec}, CLIFF~\cite{cliff}. \Oursp terms our model-based variance. We show results come from different data sources. Row 1: PDHuman test. Row 2, 3: web images. Row 4: SPEC-MTP. The number under each image represents predicted/ground-truth $f$, FoV angle, and $T_{z}$. The ground-truth $f$ and $T_{z}$ for SPEC-MTP are pseudo labels. The focal lengths here are all transformed to pixels in full image.}
  \label{fig:sota_demo}
\end{figure*}

\noindent\textbf{Results on PDHuman, SPEC-MTP, and HuMMan.}
We report PA-MPJPE, MPJPE, PVE, mIoU, and P-mIoU on these three datasets. For model-based methods, we compare with HMR~\cite{hmr}, SPEC~\cite{spec}, CLIFF~\cite{cliff}, PARE~\cite{pare}. We compare model-free methods with GraphCMR~\cite{graphcmr} and FastMETRO~\cite{fastmetro}.
From ~\cref{tab:sota}, we can see that SPEC~\cite{spec} performs poorly on these distorted datasets. This is mainly due to their wrong focal length assumption, which has negative rather than positive effects on their supervision. (Note that our re-implemented SPEC has higher performance than the official code, see in Sup. Mat.) CLIFF performs well on SPEC-MTP, while badly on PDHuman. Because their focal length assumption is about $53^{\circ}$ for 16:9 images, close to SPEC-MTP images. Although HMR-$f$ is trained with the same focal length as \Ours, it improves little compared to HMR since they have not encoded the distortion or distance feature into their network. \Ours-H48 outperforms SOTA methods on most metrics, especially the 3D ones. Some 2D re-projection metrics,  \eg mIoU and P-mIoU, of \Ours-H48 are lower than~\Oursp-R50 version. We conjecture that model-based methods have better reconstructed shapes.
Please refer to \cref{fig:sota_demo} for qualitative results.
More qualitative results and failure cases can be found in Sup. Mat.

\input{tables/sota-3dpw}
\noindent\textbf{Results on 3DPW.}
This study compares our proposed method, \Ours, with SOTA methods~\cite{hybrik, graphcmr, hmr, spin, pymaf, spec, pare, fastmetro, cliff, meshgraphormer}, including both model-based and model-free approaches. As shown in~\cref{tab:3dpw}, \Ours-R50 achieves comparable results to the SOTA method FastMETRO-R50 even without being fine-tuned on the 3DPW training set. Moreover, after fine-tuning, \Ours with both backbone structures show a significant improvement in performance. Furthermore, when using HRNet-w48, our approach outperforms all SOTA methods in all three metrics, surpassing model-based SOTA method CLIFF~\cite{cliff} and model-free SOTA method FastMETRO~\cite{fastmetro}. This superiority can be attributed to two main factors: on one hand, in training, we use ground-truth focal length and translation from 3DPW raw data to supervise the rendering of IUV images and distortion images; on the other hand, given that 96\% of the 3DPW images were captured within a distance of 1.2m to 10m, with more than half of them captured within 4m, there exist many perspective-distorted images. We provide a detailed analysis of the results for samples captured from different distances in the 3DPW dataset in the Sup. Mat.

\begin{figure}[ht]
    \centering
    \includegraphics[width=0.95\linewidth]{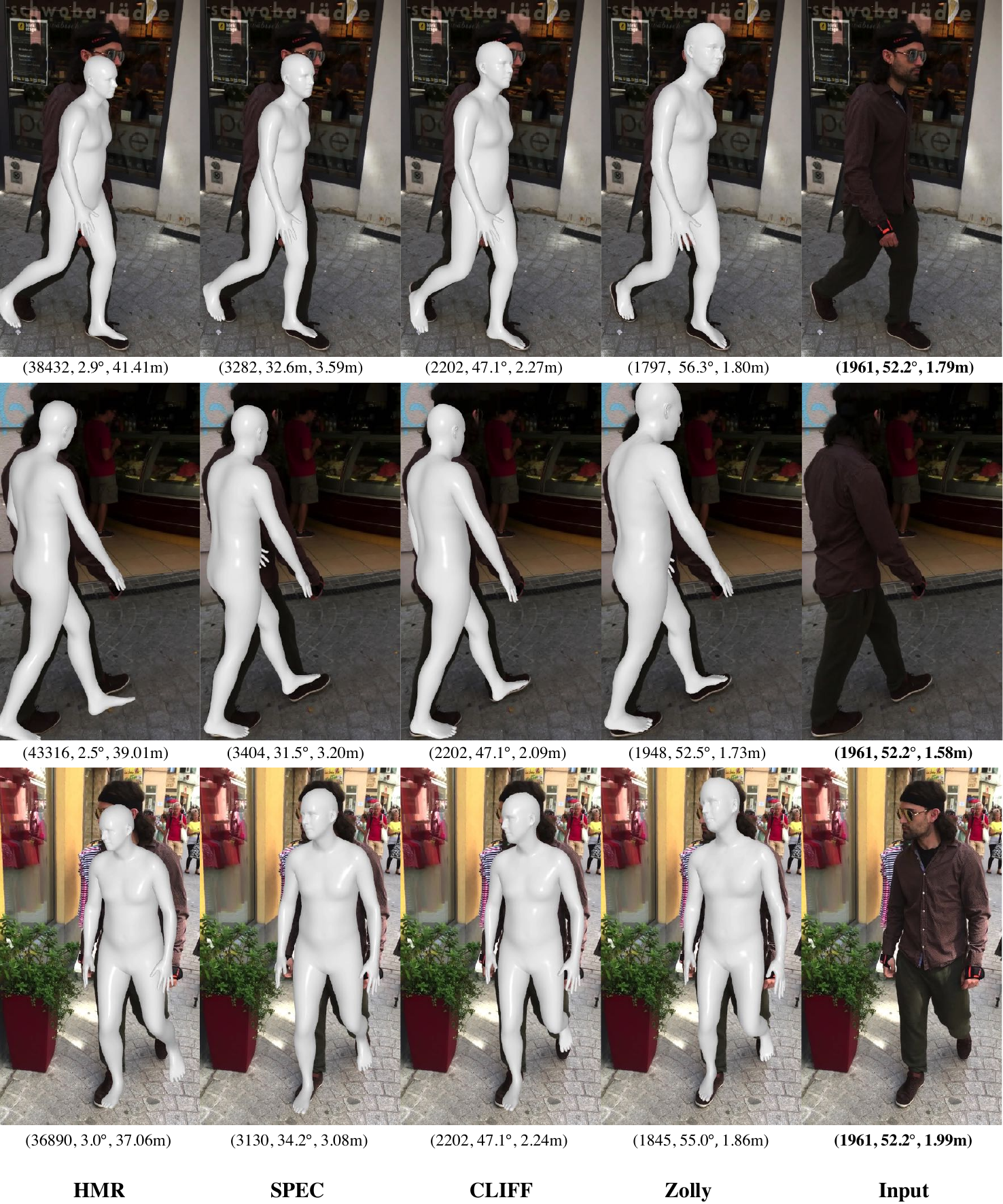}
    \caption{Qualitative results for 3DPW. \Ours achieves good alignment with the characters in the original image, but other SOTA methods have difficulty aligning images that suffer from distortion caused by overhead shots, which causes upper body dilation and lower body shrinkage. 
    The number under each image represents predicted/ground-truth $f$, FoV angle, and $T_{z}$. }
    \label{fig:demo_pw3d}
\end{figure}

\noindent\textbf{Results on Human3.6M:}
During training, we get the ground-truth focal length and translation from Human3.6M~\cite{h36m} training set for our supervision. When evaluating Human3.6M, we follow HybrIK~\cite{hybrik} by using SMPL joints as the ground truth for evaluation. As shown in~\cref{tab:h36m}, our method performs well on Human3.6M through it is not a perspective-distorted dataset. \Ours-H48 achieves the best result on the PA-MPJPE metric and achieves comparable results on the MPJPE metric. CLIFF achieves the best results on MPJPE while they also need ground-truth bounding boxes during testing.

\begin{table}[ht]
    \centering
    \scalebox{0.7}{\input{tables/supp_sota_h36m}}
    \vspace{-0.05in}
    \caption{Results of SOTA methods on Human3.6M~\cite{h36m}.
    }
    \vspace{-2ex}
    \label{tab:h36m}
\end{table}

\subsection{Ablation study}
\input{tables/ablation_3dpw}
\noindent\textbf{Ablation on training settings on the standard benchmark 3DPW.}
In Table~\ref{tab:abl_pw3d}, we present the results of our ablation study on the standard 3DPW benchmark~\cite{3dpw}, where we investigate the impact of different training settings on the performance of our method. By controlling two different variables, we show that introducing perspective-distorted datasets and fine-tuning with ground-truth focal length both lead to a slight improvement in performance. Notably, our method \Ours-H48 still outperforms the current state-of-the-art methods even without using perspective-distorted data or ground-truth focal length.


This study evaluates the effectiveness of the distortion feature and the hybrid re-projection loss function. The evaluation is conducted on the PDHuman ($\tau=3.0$), as this  exhibits the highest degree of distortion. More experimental results are provided in the Sup. Mat.

\noindent\textbf{Effect of distortion feature.}
In \cref{tab:ablation}, w/o $w(I_{d})$ terms without warp distortion image into UV space, and w/o $c(F_{d})$ terms without concatenating distortion feature to per-vertex feature. We can see that, while the mIoU and P-mIoU change a little, the 3D metrics increase significantly with the correct distortion feature. This study validates our intuition that distortion information helps the network predict more accurate vertex coordinates.

\noindent\textbf{Effect of the hybrid re-projection loss function.}
We experimented with different re-projection loss configurations and found that relying solely on weak-perspective loss significantly decreases 2D alignment. Incorporating perspective loss improved 3D metrics slightly but increased the 2D segmentation error significantly. Moreover, using per-joint distortion weight to supervise the weak-perspective camera improved the alignment of the human mesh and resulted in more accurate 3D supervision without increasing the 2D segmentation error.


We conclude that utilizing dense distortion features and an accurate camera model greatly improves the performance of our proposed method.
\input{tables/ablation_network}

%% file: tables/sota-distortion.tex
\begin{tabular}{lccccc|ccccc|ccccc}
\toprule
\multirow{2}{*}{\textbf{Methods}} & \multicolumn{5}{c}{PDHuman~($\tau=3.0$)}& \multicolumn{5}{c}{SPEC-MTP~($\tau=1.8$)}
&\multicolumn{5}{c}{HuMMan~($\tau=1.8$)}\\ \cmidrule{2-16}
&PA-MPJPE$\downarrow$ & MPJPE$\downarrow$  & PVE$\downarrow$ & mIoU$\uparrow$ &P-mIoU$\uparrow$& PA-MPJPE$\downarrow$ & MPJPE$\downarrow$ & PVE$\downarrow$ & mIoU$\uparrow$ &P-mIoU$\uparrow$& PA-MPJPE$\downarrow$ & MPJPE$\downarrow$ & PVE$\downarrow$ &  mIoU$\uparrow$  & P-mIoU$\uparrow$ \\ \midrule 

\rule{0pt}{10pt} HMR~(R50)~\cite{hmr} &62.5    &91.5   & 106.7  & 48.9 & 21.7  & 73.9& 121.4  & 145.6  & 48.8  & 16.0  & 30.2 &  43.6  & 52.6 & 65.1   & 39.5 \\

\rule{0pt}{10pt} HMR-$f$~(R50)~\cite{hmr}& 61.6 & 90.2 &105.5 & 45.2& 20.4 & 72.7 & 123.2 & 145.1 & 52.3  & 20.1  & 29.9 & 43.6& 53.4 & 62.7& 34.9  \\

\rule{0pt}{10pt} SPEC~(R50)~\cite{spec}  & 65.8&  94.9 &109.6 & 43.4   & 19.6    & 76.0 & 125.5 &144.6 &  49.9 & 18.8   & 31.4  & 44.0 & 54.2 & 51.4 & 25.6       \\

\rule{0pt}{10pt} CLIFF~(R50)~\cite{cliff} &66.2 & 99.2   &  115.2  &51.4  & 24.8   & 74.3   &  115.0 & 132.4 & 53.6  &23.7  &28.6 &42.4 &50.2  &68.8   & 44.7\\

\rule{0pt}{10pt} PARE~(H48)~\cite{pare}   & 66.3& 95.9   & 116.7 & 48.2    & 20.9   &74.2& 121.6     & 143.6 & 55.8   & 23.2    & 32.6 & 53.2& 65.5  & 66.5 & 38.3  \\


\midrule
\rule{0pt}{10pt} GraphCMR~(R50)   & 62.0  & 85.8  & 98.4  & 47.9    & 21.5  &  76.1 & 121.4    & 141.6  &  53.5 & 22.0  & 29.5 &  40.6 & 48.4  & 61.6 & 37.5   \\

\rule{0pt}{10pt} FastMETRO (H48)~\cite{fastmetro}   & 58.6 & 83.6   & 95.4 & 50.1  & 22.5   & 75.0   & 123.1  & 137.0 & 53.5  & 20.5  & 26.3  & 38.8 & 45.5  & 68.3 & 45.2    \\

\midrule
\rule{0pt}{10pt} \Oursp (R50)  & 54.3  & 80.9  & 93.9  & \textbf{54.5} &  \textbf{27.4} & 72.9 & 117.7  &  138.2 & 54.7 & 22.4 &24.4  &  36.7& 45.9 & 70.4 & \textbf{45.4}   \\
\rule{0pt}{10pt} \Ours (R50)  & 54.3&  76.4  & 87.6 & 51.4 &  24.0  & 74.0& 122.1 & 135.5  & 58.9  & 24.9  & 25.5 & 36.7 & 43.4 & 67.0 & 38.4  \\
\rule{0pt}{10pt} \Ours (H48)  & \textbf{49.9} &\textbf{70.7}  & \textbf{82.0}  & 53.0 & 26.5 &  \textbf{67.4} & \textbf{114.6} &  \textbf{126.7}& \textbf{62.3}  & \textbf{30.4} & \textbf{22.3} &\textbf{32.6} &  \textbf{40.0} & \textbf{71.2} & 45.1      \\
\bottomrule
\end{tabular}

%% file: tables/sota-3dpw.tex
\begin{table}
    \centering
     \scalebox{0.7}{
        \begin{tabular}{lccccc}
        \toprule
        \multirow{2}{*}{\textbf{Methods}} &\multirow{2}{*}{Backbone}&
        \multirow{2}{*}{w. 3DPW} 
        & \multicolumn{3}{c}{Metrics}\\ \cmidrule{4-6}
        & &  & PA-MPJPE$\downarrow$ & MPJPE$\downarrow$ & PVE$\downarrow$  \\ 
        \midrule 

        HybrIK\cite{hybrik} &  ResNet-34 & $\times$ & 48.8  & 80.0  & 94.5  \\
        HybrIK\cite{hybrik} &  ResNet-34 &$\checkmark$ & 45.0  & 74.1  & 86.5  \\
        GraphCMR~\cite{graphcmr} & ResNet-50  & $\times$  &70.2  & - &    - \\
        HMR~\cite{hmr} & ResNet-50 & $\times$ &72.6  &  116.5 & -         \\
        SPIN~\cite{spin}  &  ResNet-50 &$\times$ & 59.2     & 96.9    &116.4\\
        PyMAF~\cite{pymaf}  & ResNet-50  & $\times$ & 58.9 & 92.8  & 110.1   \\
        SPEC~\cite{spec}  &  ResNet-50  & $\checkmark$ & 52.7  & 96.4  & -  \\
        PARE~\cite{pare}  &  ResNet-50 & $\times$ &  52.3 & 82.9  & 99.7\\

        FastMETRO~\cite{fastmetro} &  ResNet-50  &$\checkmark$ & 48.3  & 77.9  & 90.6  \\
        CLIFF~\cite{cliff}  &  ResNet-50   &$\checkmark$ & 45.7  & 72.0  & 85.3   \\
        PARE~\cite{pare}  &  HRNet-w32 & $\checkmark$ & 46.5 &  74.5 &88.6 \\
        CLIFF~\cite{cliff}  &  HRNet-w48 & $\checkmark$ & 43.0 & 69.0 & 81.2        \\
        Graphormer~\cite{meshgraphormer}  &  HRNet-w64 &$\checkmark$& 45.6  & 74.7  & 87.7   \\
        FastMETRO~\cite{fastmetro} &  HRNet-w64  &$\checkmark$ & 44.6  & 73.5  & 84.1    \\

        \midrule
        $\rm\Ours^\cP$ & ResNet-50  & $\times$&    48.9      & 80.0 & 92.3\\
         \Ours & ResNet-50  & $\times$&    49.2     & 79.6 & 92.7 \\
             \Ours & ResNet-50  & $\checkmark$ & 44.1 & 72.5 & 84.3  \\

        \Ours   & HRNet-w48  & $\times$ & 47.9 & 76.2        & 89.8\\
        \Ours  & HRNet-w48  & \checkmark & \textbf{39.8}   &    \textbf{65.0} & \textbf{76.3}  \\
        \bottomrule
        \end{tabular} 
    }
    \vspace{-0.05in}
    \caption{Results of SOTA methods on 3DPW. \Oursp terms our parametric-based variant.
    `w/o PD' means trained without the proposed distorted dataset. `w/ f' means trained with ground-truth focal length if provided.
    }
    \vspace{-2ex}
    \label{tab:3dpw}
\end{table}

%% file: tables/supp_sota_h36m.tex
\begin{tabular}{lccc}
        \toprule
        \multirow{2}{*}{\textbf{Methods}} &\multirow{2}{*}{Backbone}
        & \multicolumn{2}{c}{Metrics}\\ \cmidrule{3-4}
        &  & PA-MPJPE$\downarrow$ & MPJPE$\downarrow$   \\ 
        \midrule 

        HybrIK\cite{hybrik} &  ResNet-34 & 34.5 & 54.4 \\
        HMR~\cite{hmr} & ResNet-50  & 56.8 &88.0 \\
        GraphCMR~\cite{graphcmr} & ResNet-50  &50.1 & - \\
        SPIN~\cite{spin}  &  ResNet-50 & 41.1 &  62.5 \\
        PyMAF~\cite{pymaf}  & ResNet-50  &    40.5 &57.7\\
        FastMETRO~\cite{fastmetro} &  ResNet-50  &  37.3 & 53.9\\
        CLIFF~\cite{cliff}  &  ResNet-50  &35.1  &50.5 \\
        CLIFF~\cite{cliff}  &  HRNet-w48 & 32.7 &\textbf{47.1} \\
        Graphormer~\cite{meshgraphormer}  &  HRNet-w64 & 34.5 & 51.2  \\
        FastMETRO~\cite{fastmetro} &  HRNet-w64  & 33.7 & 52.2 \\
        \midrule    
        \Oursp  & ResNet-50  & 34.7 & 54.0 \\
        \Ours  & ResNet-50  & 34.2 & 52.7 \\
        \Ours  & HRNet-w48  & \textbf{32.3} & 49.4 \\
        \bottomrule
        \end{tabular} 

%% file: tables/ablation_3dpw.tex
\begin{table}
\centering
\scalebox{0.9}{
\begin{tabular}{cccccc}
    \toprule
\multirow{2}{*}{\textbf{w/ PD}} &\multirow{2}{*}{\textbf{w/ 3DPW}} &   \multirow{2}{*}{\textbf{w/ gt $f$}} & \multicolumn{3}{c}{Metrics} \\ \cmidrule{4-6}
    &  &  & PA-MPJPE & MPJPE  & PVE  \\ 
    \midrule 

$\times$   & $\times$  &   - &   48.3     &  78.0   & 92.0  \\

$\times$   & $\checkmark$  &    $\times$  & 41.3        & 67.4     &  78.9   \\ 
$\times$  & $\checkmark$ &  $\checkmark$  & 40.9   & 67.2  & 78.4  \\ \hline

$\checkmark$  &  $\times$  &   - &      47.9  & 76.2    &  89.8\\

$\checkmark$     &  $\checkmark$  &   $\times$    & 40.9         & 66.4    & 78.3 \\ 
$\checkmark$   &    $\checkmark$   &   $\checkmark$    & \textbf{39.8}       & \textbf{65.0}       & \textbf{76.3}  \\ 

    \bottomrule
\end{tabular}}
\caption{\small Ablation study of \Ours-H48 of different training settings on 3DPW dataset. w/ PD means whether trained on perspective-distorted datasets (PDHuman, HuMMan). w/ 3DPW means whether fine-tuned on 3DPW~\cite{3dpw} dataset. w/ gt $f$ means using ground-truth focal length when fine-tuned on 3DPW.}
\label{tab:abl_pw3d}
\vspace{-10pt}
\end{table}

%% file: tables/ablation_network.tex
\begin{table}[ht]
\centering
\scalebox{0.62}{
\begin{tabular}{ccccccc}
    \toprule
\multirow{2}{*}{\textbf{Architecture}} & \multirow{2}{*}{\textbf{Loss}} & \multicolumn{5}{c}{Metrics} \\ \cmidrule{3-7}
     &  & PA-MPJPE & MPJPE  & PVE & mIoU  & P-mIoU \\ 
    \midrule 

 \rule{0pt}{10pt} \Ours w/o$\text{ }w(I_{d}), c(F_{d})$   & $\Sigma d_{J}L_{W}^{J} + L_{P}$  & 60.2  & 86.8  & 99.0 & 52.0 & 24.9     \\
 \rule{0pt}{10pt} \Ours w/o$\text{ }c(F_{d})$  & $\Sigma d_{J}L_{W}^{J} + L_{P}$  & 57.0 & 83.3 & 95.0   & 51.2 & 23.6    \\ 
 \rule{0pt}{10pt} \Ours & $L_{W}$  & 56.4  & 80.0  & 92.2 &  47.3 & 21.2     \\ 
  \rule{0pt}{10pt} \Ours & $L_{W} + L_{P}$  & 56.1 & 79.1 & 91.0 & 52.5 & 25.5    \\ 
 \midrule
 \rule{0pt}{10pt} \Ours & $\Sigma d_{J}L_{W}^{J} + L_{P}$  & 54.3 & 76.4 & 87.6  & 51.4 & 24.0    \\ 
    \bottomrule
\end{tabular}}
\caption{\small Ablation study of \Ours-H48 structure on PDHuman~($\rm \tau=3.0$). $L_{W}$ indicates weak-perspective re-projection loss. $L_{P}$ indicates perspective re-projection loss. $\sum d_{J}L_{W}^{J}$ terms dividing the per-joint distortion weight in our weak-perspective loss.}
\label{tab:ablation}
\vspace{-10pt}
\end{table}

%% file: sections/5-conclusion.tex
\section{Conclusion}
We present \Ours, the first 3DHMR method that focuses on human reconstruction from perspective distorted images. Our proposed camera model and focal length solution accurately reconstruct the human body, especially in close-view photographs. We introduce a new dataset, PDHuman, and extend two datasets containing perspective-distorted images. Our results show significant value for human mesh reconstruction in perspective-distorted images and can empower many downstream tasks, such as monocular clothed human reconstruction~\cite{selfpose, neuralbody, icon} and human motion reconstruction in live shows, vlogs, and selfie videos. Further improvements and broader applications could be explored in the future.

\noindent\textbf{Societal Impacts.} Misuse by gaming or animation companies for motion capture can lead to copyright infringement and discourage original content. Fair use advocacy and negotiation with creators can promote sustainable creativity.
\paragraph{Acknowledgement}
Taku Komura
and Wenjia Wang are partly supported by Technology Commission (Ref:ITS/319/21FP) and Research Grant Council
(Ref: 17210222), Hong Kong.

%% file: sections/6-appendix_cr.tex
\appendix


\section{Details of Perspective-distorted Datasets.}
\subsection{PDHuman}
\label{chapter:pdhuman-syn}
Our pipeline is inspired by recent works on synthetic data~\cite{Patel:CVPR:2021:AGORA,hspace,varol17_surreal}. 
A photogrammetry-scanned human model with a unique body pose will be rendered with a random viewpoint in an HDRi background. The detailed statistics of PDHuman are illustrated in \cref{tab:pdhuman_stat}.

\begin{table}[htp]
    \centering
    \scalebox{0.75}{
    \input{tables/supp_pdhuman_info}}
    \caption{Statistical information of PDHuman. $\tau$ denotes maximum diisortion scale in the main text.}
    \label{tab:pdhuman_stat}
    \vspace{-5pt}
\end{table}

\noindent\textbf{Human model.}
We use a corpus of $630$ photogrammetry-scanned human models from Renderpeople~\cite{renderpeople},
with well-fitted SMPL parameters. 
Initially, the body pose is sampled from a collection of high-quality motion sequences obtained from
Mixamo~\cite{mixamo}.
The the pose is converted to SMPL skeleton using a re-targeting approach.
Finally, we use a SGD optimizer to optimize the chamfer distance between the SMPL vertices and RenderPeople vertices to refine the pose and shape parameter.

\noindent\textbf{Camera.}
In order to simulate a wide range of real-world scenarios, a perspective camera is randomly sampled with a focal length that spans from 7mm to 102mm. The corresponding FoV angle is from 10\degree to 140\degree.
The human mesh is then positioned at the center of a sphere, whose radius is chosen randomly and dependent on the camera's focal length.
The camera, facing the center of the sphere, is then placed on the surface of the sphere with a randomized elevation and azimuth angle.

\noindent\textbf{Rendering pipeline.} 
To increase the diversity of data, each frame contains ambient lighting calculated by path tracing in Blender~\cite{blender} and diverse background generated by HDRi images from PolyHaven~\cite{polyhaven}. The size of all rendered images is $512\times512$.

%

\begin{figure}
    \centering
    \includegraphics[width=1.0\linewidth]{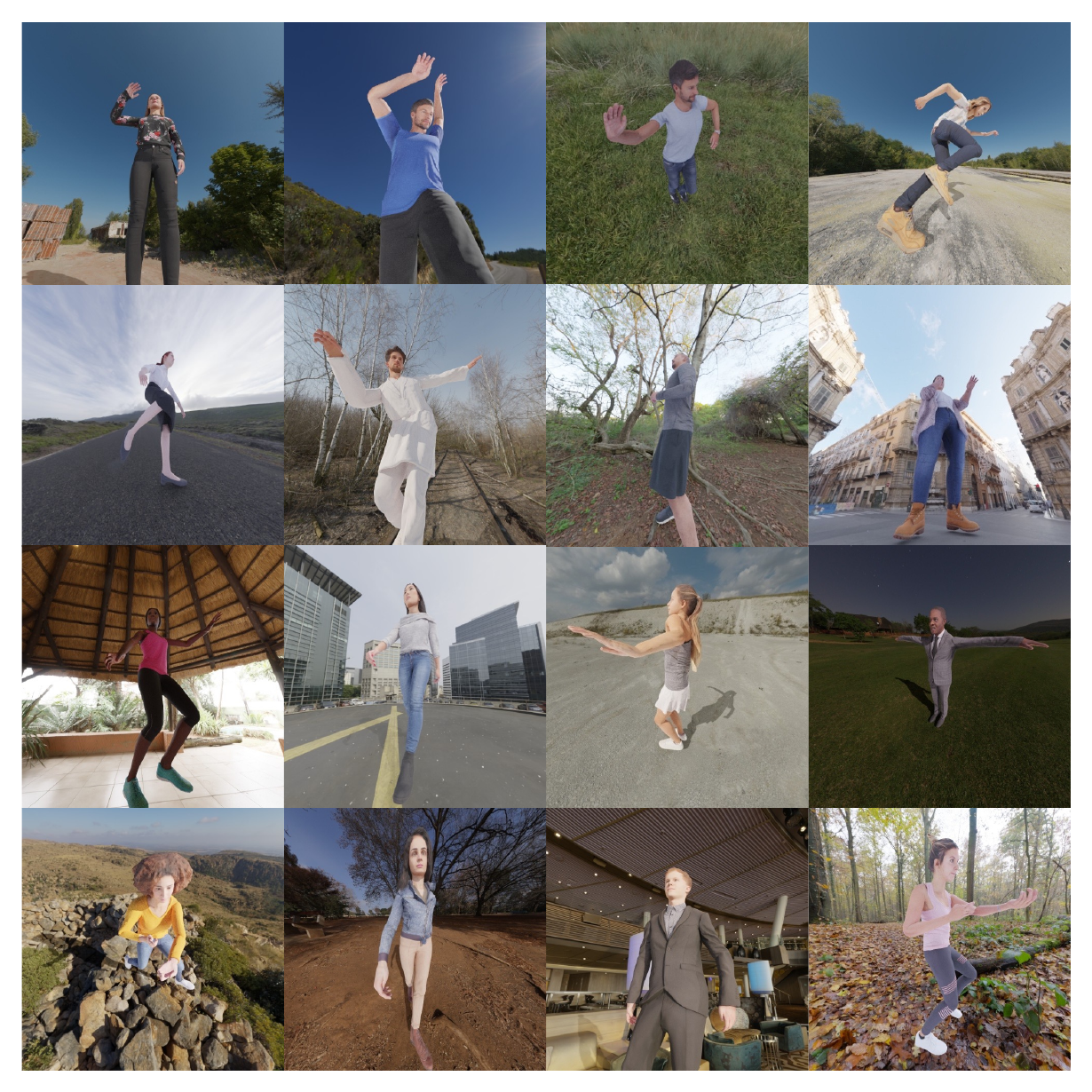}
    \vspace{-5pt}
    \caption{More examples from PDHuman dataset.}
    \label{fig:pdhuman_demo}
    \vspace{-5pt}
\end{figure}

\subsection{SPEC-MTP}
\label{chapter:spec-mtp}

\begin{table}
    \centering
    \scalebox{0.8}{\input{tables/supp_spec_mtp_info}}
    \caption{Statistical information of SPEC-MTP~\cite{spec}. $\tau$ denotes the maximum distortion scale in the main text.}
    \label{tab:spec_mtp_stat}
    \vspace{-5pt}
\end{table}

SPEC-MTP~\cite{spec} is a real-world image dataset with calibrated focal lengths and well-fitted SMPL parameters (including $\theta$, $\beta$), and translation. The images were taken at relatively close-up distances, leading to noticeable perspective distortion in the limbs and torsos of subjects. We use it as one of the evaluation datasets for our task. The detailed statistics of SPEC-MTP are illustrated in \cref{tab:spec_mtp_stat}.


\subsection{HuMMan}
The HuMMan dataset, as proposed in \cite{humman}, is a real-world image dataset that utilizes 10 calibrated RGBD kinematic cameras to capture shots for each frame. From these shots, segmented point clouds are extracted from depth images, resulting in a comprehensive dataset. The SMPL parameters were registered upon triangulated 3D keypoints and point clouds, providing ground-truth data that is highly useful for mocap tasks. We reshape all the images to $360\times 640$ pixels. The detailed statistics of HuMMan are illustrated in~\cref{tab:humman_stat}.


\begin{table}
    \centering
    \scalebox{0.75}{
\input{tables/supp_humman_info}}
\caption{Statistical information of HuMMan~\cite{humman}. $\tau$ denotes the maximum distortion scale in the main text.}
\label{tab:humman_stat}
\vspace{-5pt}
\end{table}

\section{Analysis of 3DPW dataset}
We divide the 3DPW dataset into three protocols based on the maximum distortion scale $\tau$ in~\cref{tab:supp_pw3d_info}. We report the results of our re-implemented HMR-R50~\cite{hmr} and \Ours-H48 on each protocol in \cref{tab:supp_pw3d}.
The experiments indicate that \Ours outperforms HMR-R50, and this improvement is more pronounced as the distortion scale increases.
This observation serves as compelling evidence that \Ours's success on 3DPW can be primarily attributed to its superior performance on distorted images.
%
%

\begin{table}[ht]
    \centering
    \scalebox{0.75}{\input{tables/supp_3dpw_info}}
    \caption{3 protocols of 3DPW divided by $\tau$. The larger the value of $\tau$, the greater the degree of distortion.}
    \label{tab:supp_pw3d_info}
\end{table}

\begin{table}
    \centering
    \scalebox{0.75}{
    \input{tables/supp_pw3d}}
    \caption{Reults of our re-implemented HMR~\cite{hmr} and \Ours-H48 on different protocols of 3DPW. Mainly for showing the correlation between performance improvement and distortion.}
    \label{tab:supp_pw3d}
\end{table}


\section{Details of Model-based Variant of \Ours}
\begin{figure*}[ht]
    \centering
    \includegraphics[width=0.95\linewidth]{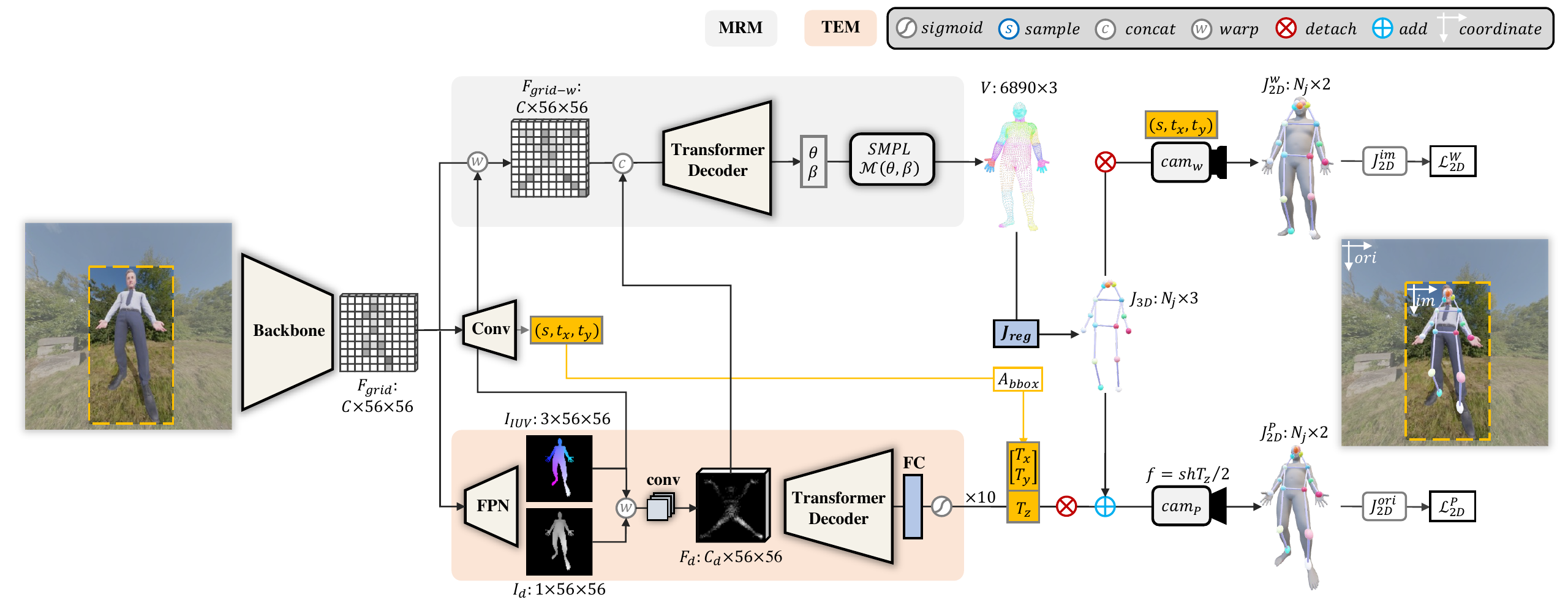}

    \caption{\textbf{\Oursp pipeline overview}. Compared to \Ours, the main difference is the reformulated mesh reconstruction module.}
    \label{fig:pipeline_p}
\end{figure*}
As shown in~\cref{fig:pipeline_p}, we introduce a model-based \Oursp by changing the mesh reconstruction module. Different from \Ours, we regress SMPL parameters rather than 3D vertex coordinates through a transformer decoder in \Oursp. We warp the grid Feature $F_{grid}$ into UV space ($F_{grid-w}$) via $I_{IUV}$ to eliminate the spatial distortion of each part of the features, then concatenate the warped distortion feature $F_{grid-w}$ and regress SMPL parameters from it. We represent the 24 rotations of joints $\theta$ and body shape parameters $\beta$ as 25 learnable tokens. The translation estimation module and supervision are exactly the same as \Ours.

\section{More about Cameras}
\noindent\textbf{Affine Transformation.}
Our affine transformation of translation is the same as SPEC~\cite{spec}.
$T_{x}, T_{y}$ and $t_{x},t_{y}$ should satisfy the following equation for every $x, y$ by connecting the re-projected coordinates in the cropped image coordinate system and original image coordinate system in screen space.
\begin{equation}
\small
\begin{bmatrix}  
\frac{w}{2}\left[ s_{x}(x+t_{x})) +1\right]+c_{x}-\frac{w}{2}\\[6pt]
\frac{h}{2}\left[ s_{y}(y+t_{y})) +1\right]+c_{y}-\frac{h}{2}
  \end{bmatrix}=
\begin{bmatrix}\frac{W}{2}\left[S_{W}(x+T_{x}) + 1\right]\\[6pt]
\frac{H}{2}\left[ S_{H}(y+T_{y}) + 1\right] \end{bmatrix}.
\end{equation}
where $S_{W}=s_{x}/(\frac{W}{w})$, $S_{H}=s_{y}/(\frac{H}{h})$, so we can get the transform by:

\begin{equation}
\small
\begin{bmatrix}
T_{x}\\ 
T_{y}\\ 
1
\end{bmatrix}=\begin{bmatrix}
 1& 0 & (2c_{x}-W)/ws_{x}\\ 
 &1  & (2c_{y}-H)/hs_{y}\\ 
 &  & 1
\end{bmatrix}\begin{bmatrix}
t_{x}\\ 
t_{y}\\ 
1
\end{bmatrix}
\end{equation}
$c_{x}, c_{y}$ terms the bounding-box center coordinate in original image. $h, w$ terms the cropped image size, where $H, W$ terms original image size. During training, we will expand every bounding box of the human body to a square and resize the cropped image to $224\times224$ pixels.

\noindent\textbf{Analysis of dolly zoom.}
The dolly zoom is an optical effect performed in-camera, whereby the camera moves towards or away from a subject while simultaneously zooming in the opposite direction.
It was first proposed in the film JAW~\cite{jaws1975}. In this section, we simulate the effect on CMU-MOSH~\cite{mosh} data.
First, we get all the vertices in CMU-MOSH~\cite{mosh} by feeding the SMPL~\cite{smpl} parameters to the body model. 
We further obtain 3D joints by multiplying the joint regressor matrix to the vertices. 
%
We then apply weak-perspective camera parameters $(s, t_{x}, t_{y})$ while adjusting the distance to approximate the human body's location and size, producing increased distortion as the camera approaches.
The weak-perspectively projected 2D joints are $s(x+t_{x}, y+t_{y})$, where $x, y$ are the corresponding 3D joint coordinates. We set the image height to 224 pixels and re-project the 2D joints and compare the error between the weak-perspective and perspective projection results.
As shown in \cref{tab:supp_dolly}, when the subject is located over 4 meters away, the re-projected error is only 1.76 pixels, which is negligible on a 224-pixel image. When the subject is further than 8 meters, the error is less than 1 pixel, indicating non-distortion of the images. 


 \begin{table}
    \centering
    \scalebox{0.7}{\input{tables/supp_dollyzoom}}
    \caption{Distortion and re-projection error caused by distance in CMU-MOSH~\cite{mosh}. The re-projected error is measured in pixels.}
    \label{tab:supp_dolly}
    \vspace{-10pt}
\end{table}

\begin{figure}[t]
    \centering
    \includegraphics[width=0.9\linewidth]{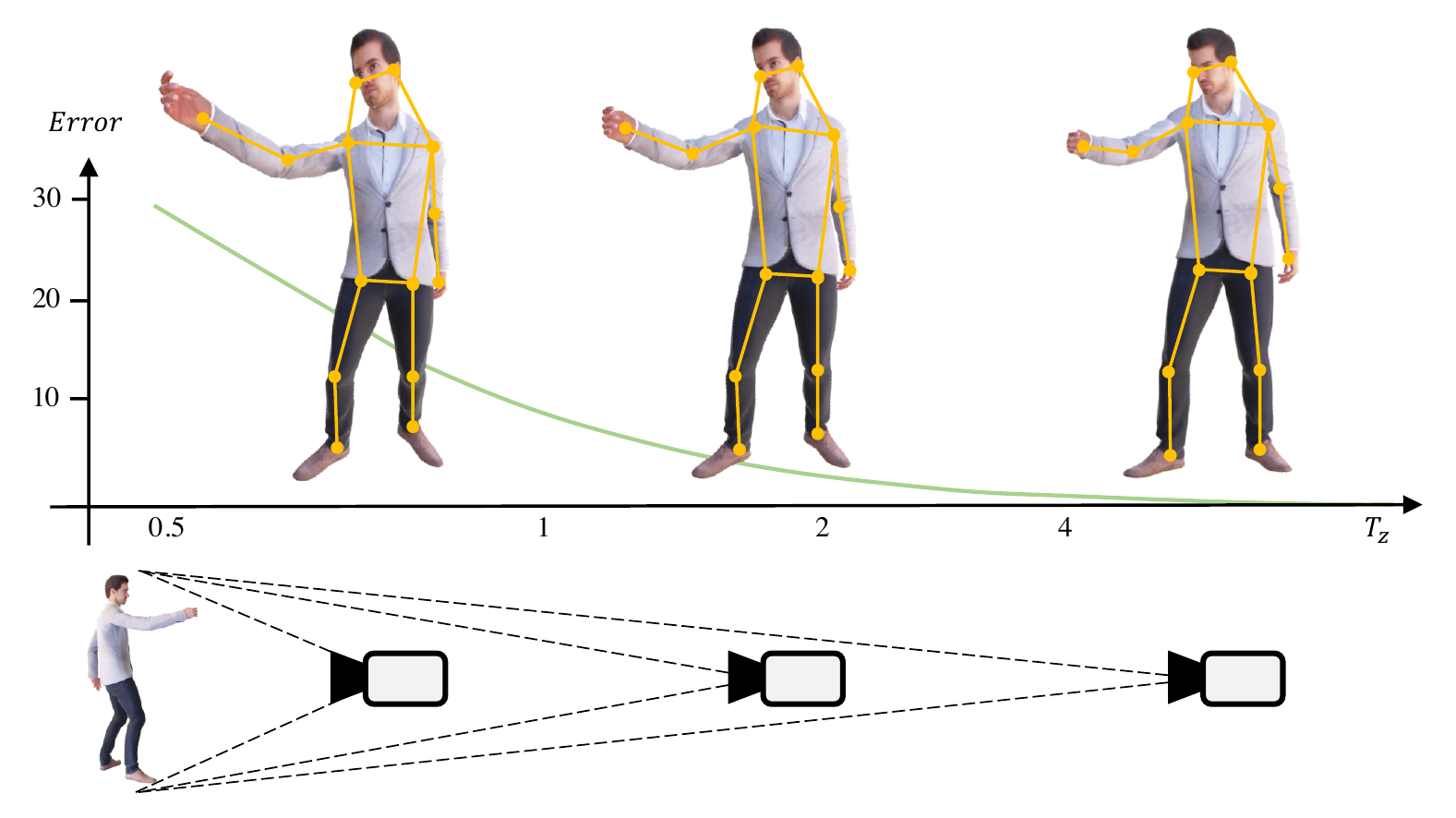}
    \caption{Distortion and re-projection error caused by distance. The vertical axis is measured in pixels, and the horizontal axis is measured in meters.}
    \label{fig:supp_dolly}
    \vspace{-5pt}
\end{figure}  

\section{More quantitative results}

\noindent\textbf{Full results on PDHuman:} As shown in~\cref{tab:sota_supp_pdhuman} and \cref{tab:sota_supp_pdhuman2}, we report results on all 5 protocols in the PDHuman test dataset. Our proposed methods, \Ours(H48) and \Oursp (R50) outperform the other methods in all metrics by a large margin. 

\noindent\textbf{Full results on SPEC-MTP:} As illustrated in~\cref{tab:sota_supp_specmtp}, we report the results of all the 3 protocols in SPEC-MTP dataset. In this real-world dataset, \Ours (H48) largely outperforms other methods in all metrics. In column $\tau=1.0$, Note that our re-implemented SPEC$*$ achieves higher performance than the official implementation.
 
\noindent\textbf{Full results on HuMMan:} As shown in \cref{tab:sota_sup_humman}, \Ours (H48) largely outperforms other methods in all metrics. By contrast, although CLIFF~\cite{cliff} performs comparably well on the HuMMan dataset, it demonstrates poor performance on the PDHuman dataset. We conjecture the focal length assumption of CLIFF is suitable for datasets captured by fixed and similar camera settings, \eg HuMMan dataset, while not valid for the PDHuman dataset with varied camera settings.

\begin{table*}[hb]
    \centering
    \scalebox{0.61}{\input{tables/supp_pdhuman}}
    
    \caption{Results of SOTA methods on PDHuman ($\tau=3.0$, $\tau=2.6$, $\tau=2.2$ protocols). HMR-$f$ terms HMR~\cite{hmr} model trained with same focal length as \Ours.}
    \label{tab:sota_supp_pdhuman}
\end{table*}

\begin{table*}[h]
    \centering
    \scalebox{0.61}{\input{tables/supp_pdhuman2}}
    
    \caption{Results of SOTA methods on PDHuman ($\tau=1.8$, $\tau=1.4$ protocols).}
    \label{tab:sota_supp_pdhuman2}
\end{table*}

\begin{table*}[h]
    \centering
    \scalebox{0.61}{\input{tables/supp_spec_mtp}}
    
    \caption{Results of SOTA methods on SPEC-MTP ($\tau=1.8$, $\tau=1.4$, $\tau=1.0$ protocols). SPEC-MTP~($\tau=1.0$) indicates the original SPEC-MTP~\cite{spec} dataset. SPEC~* terms the results reported in SPEC~\cite{spec}.}
    \label{tab:sota_supp_specmtp}
\end{table*}


\begin{table*}[h]
    \centering
    \scalebox{0.61}{\input{tables/supp_humman}}
    
    \caption{Results of SOTA methods on HuMMan ($\tau=1.8$, $\tau=1.4$, $\tau=1.0$ protocols).}
    \label{tab:sota_sup_humman}
\end{table*}

\section{Qualitative results.}

\begin{figure}[ht]
    \centering
    \includegraphics[width=1.\linewidth]{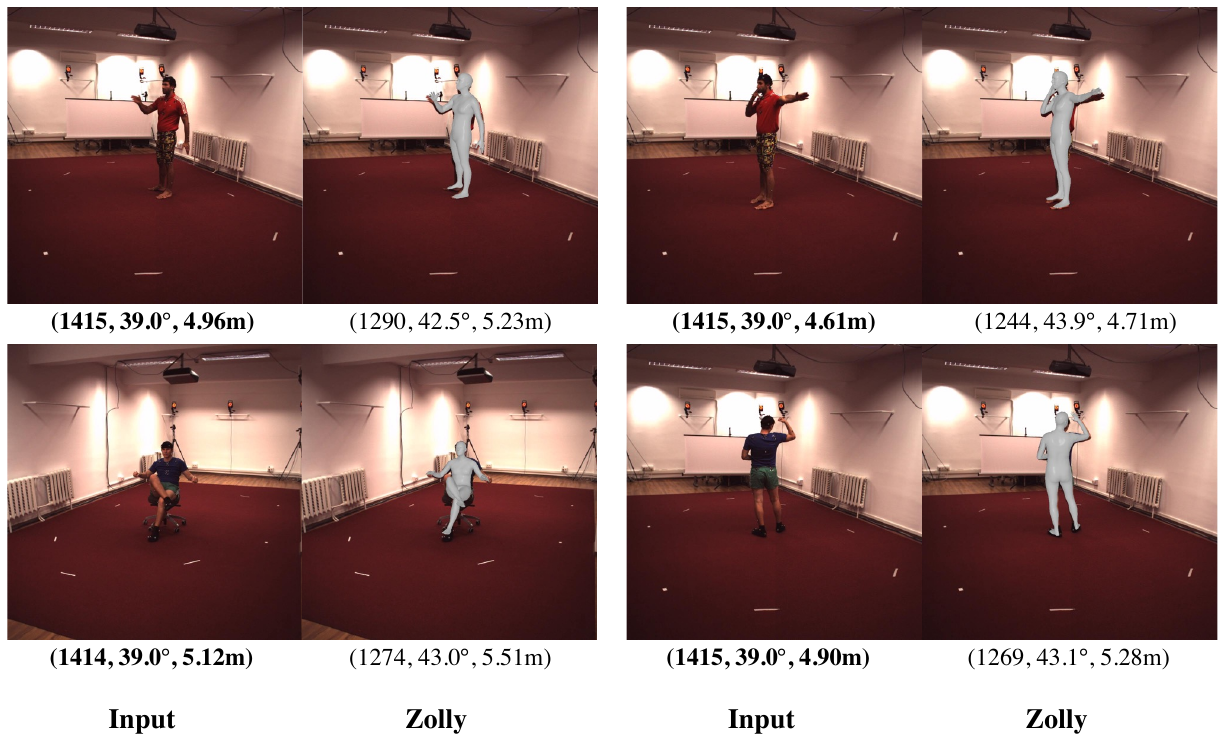}
    \caption{Qualitative results on Human3.6M dataset. The number under each image represents predicted/ground-truth focal length $f$, FoV angle, and z-axis translation $T_{z}$. Our method could predict an approximate translation for non-distorted images as well.}
    \label{fig:h36m}
    \vspace{-10pt}
\end{figure}

\noindent\textbf{Qualitative results on Human3.6M~\cite{h36m} dataset.}
We show qualitative results of \Ours on Human3.6M dataset in \cref{fig:h36m}

\begin{figure}[h]
    \centering
    \includegraphics[width=1\linewidth]{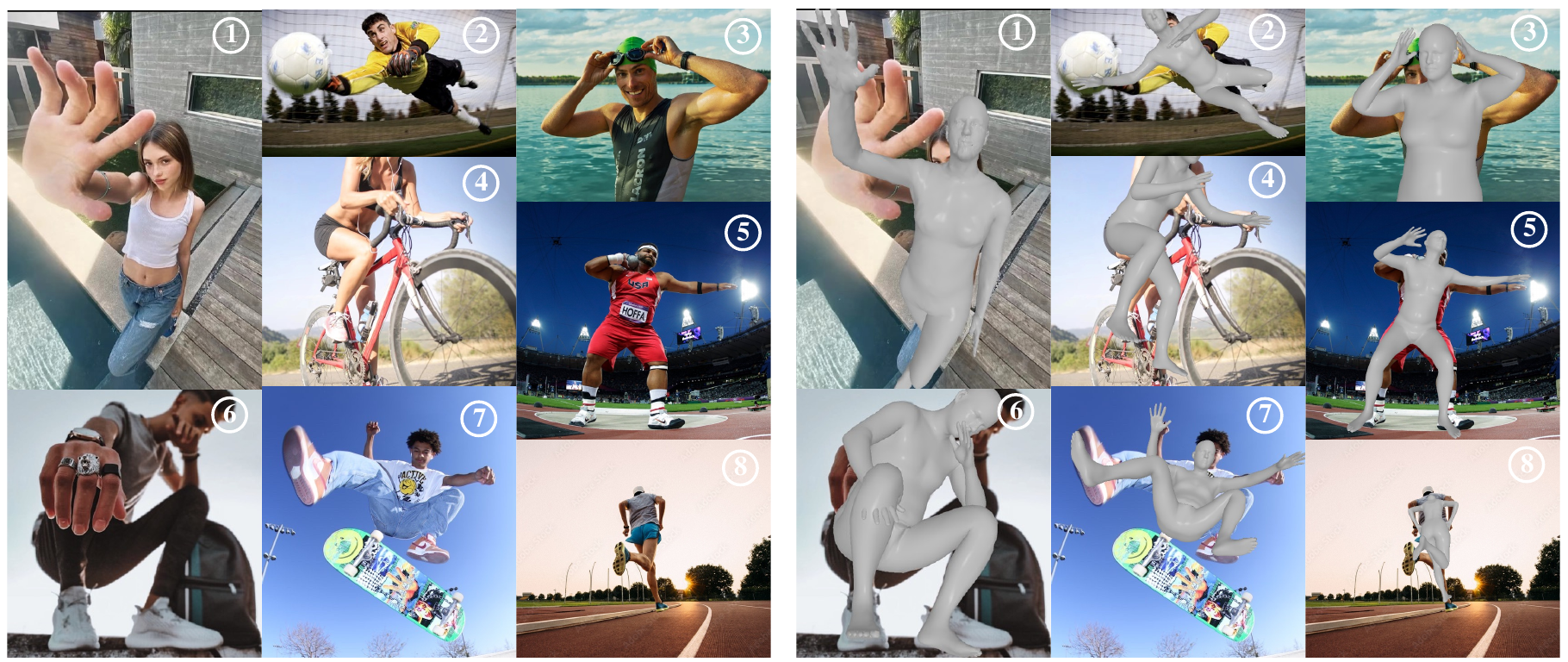}
    \caption{Failure cases. The left part is input, and the right part is our prediction.}
    \label{fig:failure}
    \vspace{-10pt}
\end{figure}

\noindent\textbf{Failure cases.}
Although our methodology is generally effective, it has trouble under certain extreme circumstances. As demonstrated in~\cref{fig:failure}, due to the lack of training data containing characters with large hands ((1), (2), and (6)), and large feet ((7) and (8)), \Ours produce sub-optimal results on such images. 
Similarly, our approach may not perform well on characters with exceptional body shapes, as exemplified by (2), (3), and (7), where the athletes have muscular bodies. Additionally, it is difficult for \Ours to reconstruct self-occluded human bodies, as depicted in (4). We are actively exploring strategies to address these limitations and improve the robustness of our methodology.

\begin{figure*}
    \centering
    \includegraphics[width=1.0\linewidth]{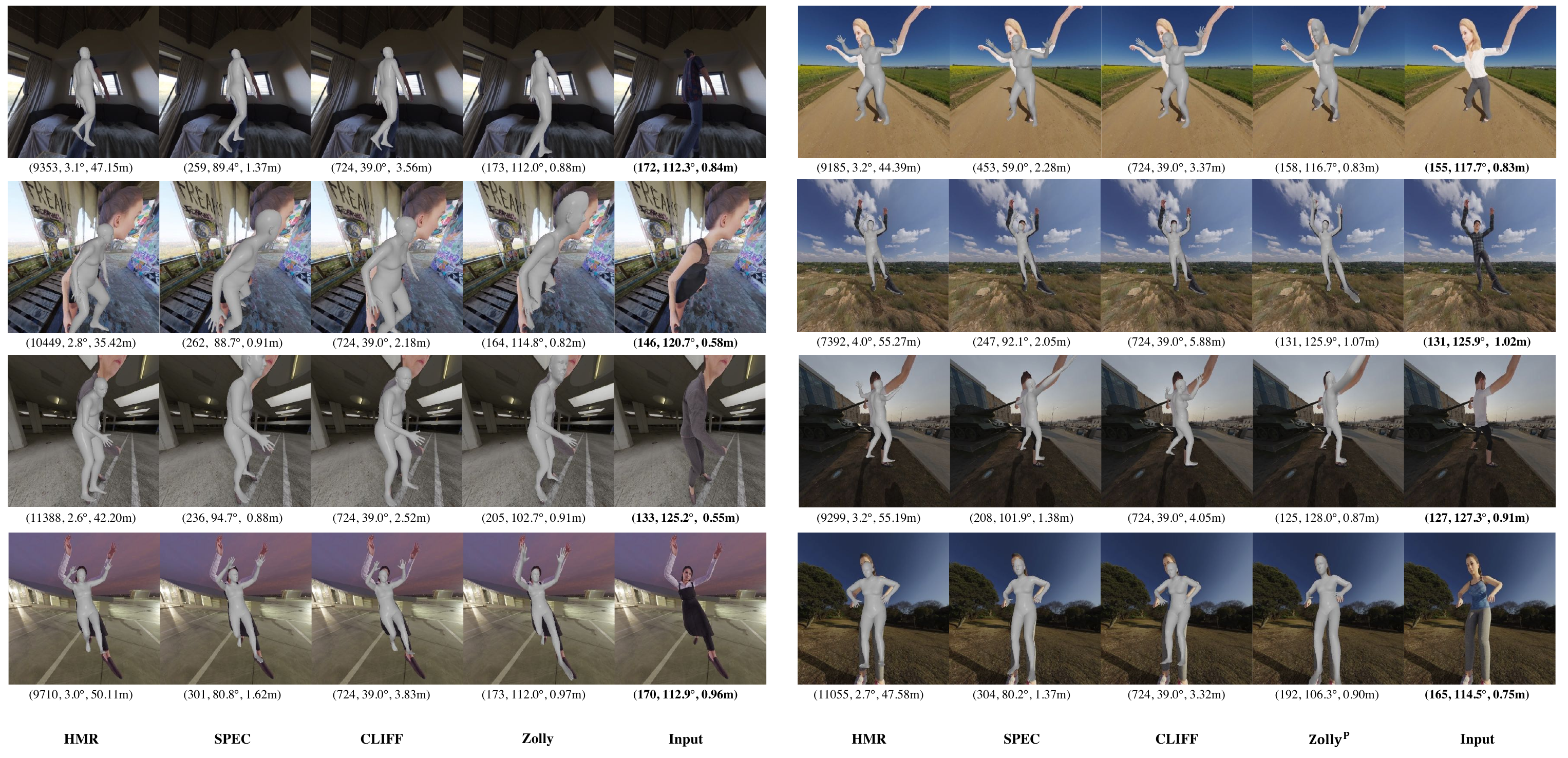}
    \caption{Qualitative results on PDHuman dataset. The number under each image represents predicted/ground-truth focal length $f$, FoV angle, and z-axis translation $T_{z}$. }
    \label{fig:supp_pdhuman_sota}
\end{figure*}

\begin{figure*}[htp]
    \includegraphics[width=1.0\linewidth]{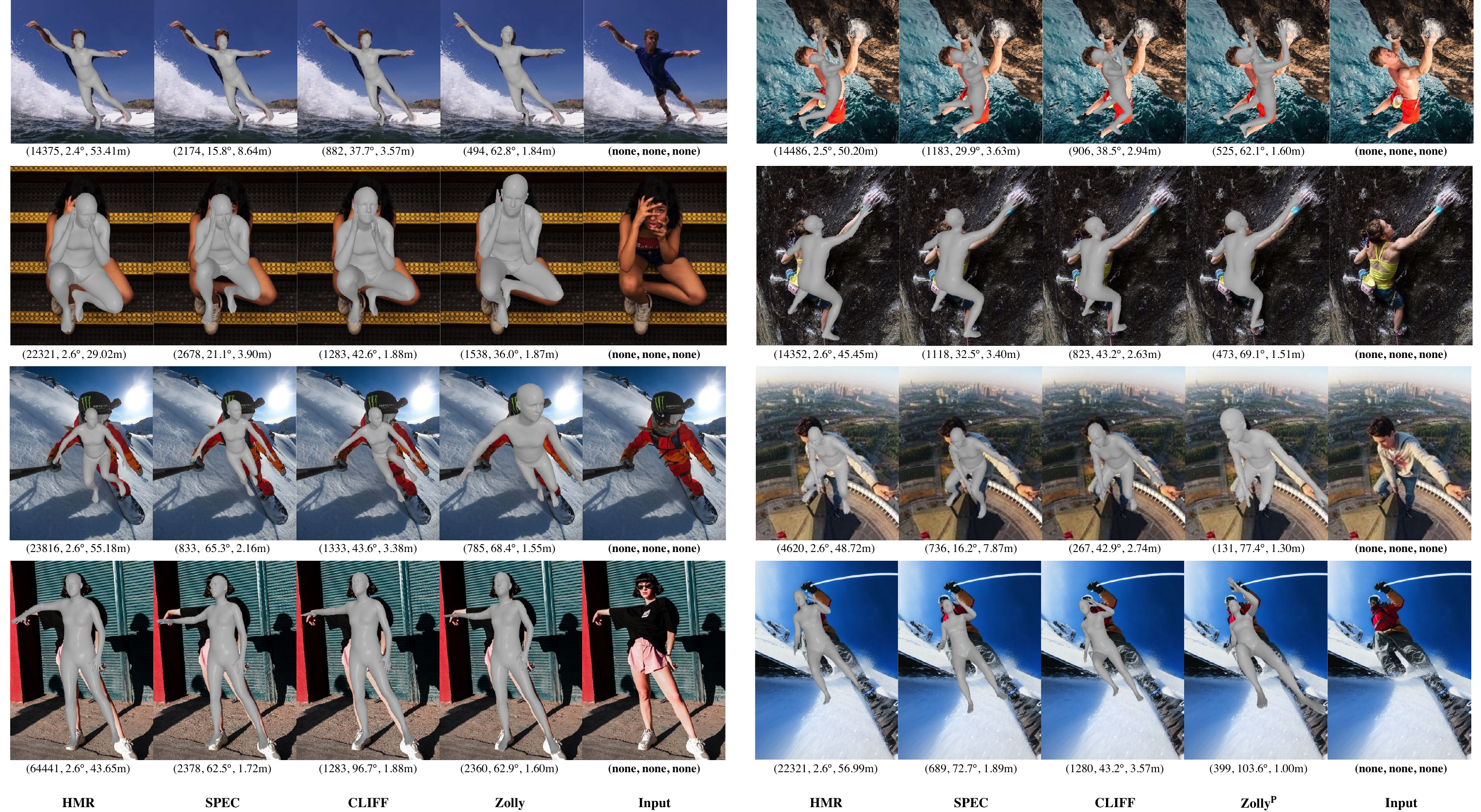}
\caption{Qualitative results on in-the-wild images. The number under each image represents the predicted focal length $f$, FoV angle, and z-axis translation $T_{z}$. Images are collected from \url{https://pexels.com} and \url{https://yandex.com}.
}
\label{fig:supp_real_sota}
\end{figure*}

\begin{figure*}
    \includegraphics[width=1.0\linewidth]{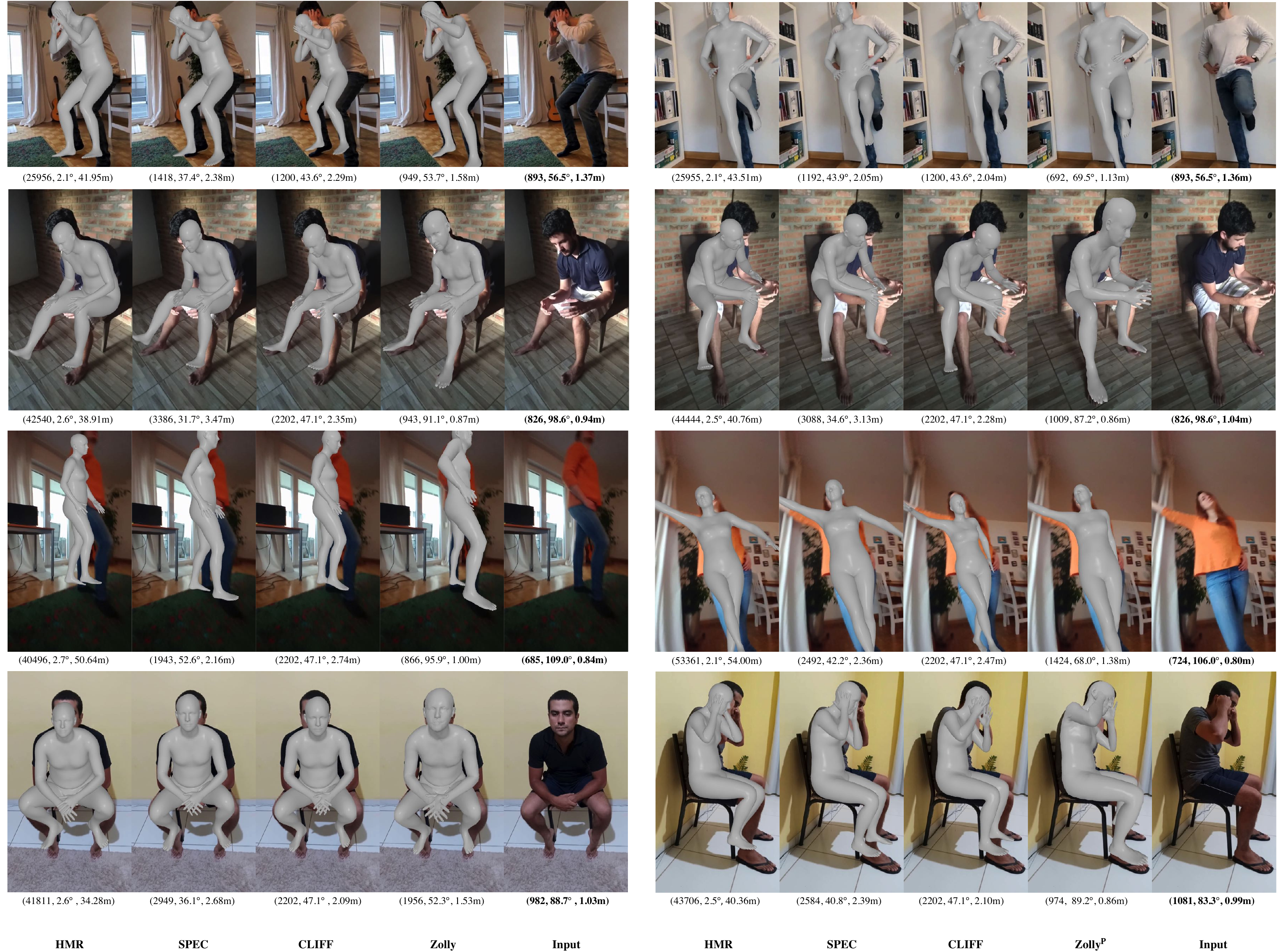}
    \caption{Qualitative results on SPEC-MTP dataset. The number under each image represents predicted/ground-truth focal length $f$, FoV angle, and z-axis translation $T_{z}$. The ground-truth $T_{z}$ and focal length $f$ for SPEC-MTP are pseudo labels.}
    \label{fig:supp_specmtp_sota}
\end{figure*}

\noindent\textbf{More qualitative results on distorted images.}
We show more qualitative results of \Ours comparing with SOTA methods for perspective-distorted images on PDHuman (\cref{fig:supp_pdhuman_sota}), Web images (\cref{fig:supp_real_sota}), and SPEC-MTP (\cref{fig:supp_specmtp_sota}).

\clearpage

%% file: tables/supp_pdhuman_info.tex
\begin{tabular}{l|p{25pt}<{\raggedright}p{25pt}<{\raggedright}p{25pt}<{\raggedright}p{25pt}<{\raggedright}p{25pt}<{\raggedright}p{25pt}<{\raggedright}}
\toprule
Protocol             & Train & 5    & 4    & 3    & 2     & 1     \\ \midrule
Number             &  126198 & 2821 & 4166 & 6601 & 12225 & 27448 \\ 
$\tau$ & 1.0 & 3.0  & 2.6  & 2.2  & 1.8   & 1.4   \\

Mean $f$ &  245    &   174  & 176     &     180 & 191  &    230   \\ 
Mean FoV       &  $\rm 98.6^\circ$       &     $\rm 113.7^\circ$  &  $\rm 112.0^\circ$     &  $\rm 112.0^\circ$    &    $\rm 110.0^\circ$     &   $\rm 102.0^\circ$    \\ 
Mean $T_{z}$      &  1.3  &    0.8  &    0.8  &   0.8   &    0.9   &    1.1   \\

\bottomrule
\end{tabular}

%% file: tables/supp_spec_mtp_info.tex
\begin{tabular}{l|p{50pt}<{\raggedright}p{50pt}<{\raggedright}p{50pt}<{\raggedright}}
\toprule
Protocol              & 3    & 2     & 1     \\ \midrule
Number                & 713 &2609  & 6083  \\ 
$\tau$    & 1.8   & 1.4   & 1.0\\ 
Mean $f$  & 935     &     976  &     1114  \\ 
Mean FoV     &     $69.3^\circ$      &   $67.7^\circ$    &       $62.4^\circ$      \\ 
Mean $T_{z}$       &1.1        & 1.1 &     1.4  \\ \bottomrule
\end{tabular}

%% file: tables/supp_humman_info.tex
\begin{tabular}{l|p{40pt}<{\raggedright}p{40pt}<{\raggedright}p{40pt}<{\raggedright}p{40pt}<{\raggedright}}
\toprule
Protocol   &Train            & 3    & 2     & 1     \\ \midrule
Number                & 84170  & 926 & 2696 & 6550 \\ 
$\tau$    & 1.0 & 1.8   & 1.4   & 1.0\\ 
Mean $f$ &    318  & 318       &   318&  318  \\ 
Mean FoV              &  127    &127 &  127     & 127      \\ 
Mean $T_{z}$           &  1.9    &  1.9 & 1.9      &  1.9     \\ \bottomrule
\end{tabular}

%% file: tables/supp_3dpw_info.tex
\begin{tabular}{l|p{45pt}<{\raggedright}p{45pt}<{\raggedright}p{45pt}<{\raggedright}}\toprule
 Protocol  &            1  &   2         &      3      \\\midrule
Number                &     35115         &            19016        &   4657           \\

$\tau$ &        1.0      &      1.08      &    1.16              \\
mean $f$                & 1966             &      1966       &      1966           \\
mean FoV              &  $52^\circ$            & $52^\circ$                 &        $52^\circ$ \\
mean $T_{z}$    & 4.6 & 3.5 & 2.9\\
\bottomrule       
\end{tabular}

%% file: tables/supp_pw3d.tex
\begin{tabular}{l|p{55pt}<{\centering}p{55pt}<{\centering}p{55pt}<{\centering}}
\toprule
\multirow{2}{*}{\textbf{Protocol/Method}} & \multicolumn{3}{c}{3DPW Test}  \\ \cmidrule{2-4}
&PA-MPJPE$\downarrow$ & MPJPE$\downarrow$  & PVE$\downarrow$ \\ \midrule 

\rule{0pt}{10pt} p1(HMR-R50)  &  50.2&  80.9  & 94.5 \\
\rule{0pt}{10pt} p1(\Ours-H48)  &39.8  & 65.0 &76.3 \\
\rule{0pt}{10pt} Improvement  &+10.4   & +15.9& +18.2\\\midrule

\rule{0pt}{10pt} p2(HMR-R50)  &51.3   &82.3  &96.2\\
\rule{0pt}{10pt} p2(\Ours-H48)  & 39.9  & 64.8 & 76.8 \\
\rule{0pt}{10pt} Improvement  & +11.4 & +17.5 &+19.4\\\midrule

\rule{0pt}{10pt} p3(HMR-R50)  & 58.3 &93.9  &107.8  \\
\rule{0pt}{10pt} p3(\Ours-H48)  &44.7  & 71.7 & 84.6\\
\rule{0pt}{10pt} Improvement  & + 13.6  & +22.2   & +23.2 \\



\bottomrule

\end{tabular}

%% file: tables/supp_dollyzoom.tex
\begin{tabular}{l|p{20pt}
<{\raggedright}p{20pt}<{\raggedright}p{20pt}<{\raggedright}p{20pt}<{\raggedright}p{20pt}<{\raggedright}p{20pt}<{\raggedright}p{20pt}<{\raggedright}p{20pt}<{\raggedright}p{20pt}<{\raggedright}p{20pt}<{\raggedright}}\toprule
$T_{z}$  &      0.5 &0.75&1.0 & 2.0 & 4.0 & 8.0 & 12.0 & 16.0 &20       \\\midrule
$\tau$  &    3.06&1.69&1.41&1.16&1.07&1.04&1.02&1.02&1.01          \\
$Error$  & 30.56&10.46&7.38&3.54&1.76&0.88&0.59&0.44&0.35              \\

\bottomrule       
\end{tabular}

%% file: tables/supp_pdhuman.tex
\begin{tabular}{lccccc|ccccc|ccccc}
\toprule
\multirow{2}{*}{\textbf{Methods}} & \multicolumn{5}{c}{PDHuman~($\tau=3.0$)}  & \multicolumn{5}{c}{PDHuman~($\tau=2.6$)}
&\multicolumn{5}{c}{PDHuman~($\tau=2.2$)} \\ \cmidrule{2-16}
&PA-MPJPE$\downarrow$ & MPJPE$\downarrow$  & PVE$\downarrow$ & mIoU$\uparrow$ &P-mIoU$\uparrow$& PA-MPJPE$\downarrow$ & MPJPE$\downarrow$ & PVE$\downarrow$ & mIoU$\uparrow$ &P-mIoU$\uparrow$& PA-MPJPE$\downarrow$ & MPJPE$\downarrow$ & PVE$\downarrow$ &  mIoU$\uparrow$  & P-mIoU$\uparrow$ \\ \midrule 

\rule{0pt}{10pt} HMR~(R50)~\cite{hmr}  & 62.5 & 91.5 & 106.6 & 48.9& 21.7 & 59.9 &87.8 & 102.4 & 50.0 & 22.5 & 57.4 & 84.0 & 98.1 & 51.4 & 23.6 \\

\rule{0pt}{10pt} HMR-$f$~(R50)~\cite{hmr}& 61.6 & 90.2 & 105.5 & 45.2 & 20.4 & 59.2 & 86.6 & 101.3 & 46.5 & 21.4 &56.8 & 82.9 & 97.2 & 48.1 & 22.7  \\

\rule{0pt}{10pt} SPEC~(R50)~\cite{spec}  &  65.8 & 94.9 & 109.6 & 43.4 & 19.6 & 63.2 & 91.5 & 105.8 &43.3 & 19.5 &60.6 & 87.3 & 101.3 & 42.2 & 18.7   \\

\rule{0pt}{10pt} CLIFF~(R50)~\cite{cliff} & 66.2 & 99.2 & 115.2 & 51.4 & 24.8 & 63.4 & 94.4 & 109.8 & 52.7 & 25.9 & 60.6 & 89.6 & 104.3 & 54.2 & 27.1 \\

\rule{0pt}{10pt} PARE~(H48)~\cite{pare} & 66.3 & 95.9 & 116.7 & 48.2 & 20.9 &63.6& 92.3 & 112.7 & 49.3 & 21.7 & 60.6 & 88.7 & 108.6 & 50.7 & 22.7   \\

\midrule
\rule{0pt}{10pt} GraphCMR~(R50)   &  62.1 & 85.8 & 98.4 & 47.9 & 21.5 & 59.5 & 82.6 & 94.8 & 49.1 & 22.4 & 56.8 & 78.8 & 90.4 & 50.5 & 23.6\\

\rule{0pt}{10pt} FastMetro(H48)~\cite{fastmetro}   &58.6 & 83.6 & 95.4 & 50.1 & 22.5 & 55.8 & 79.9 & 91.4 & 51.4 & 23.5  & 53.1 & 75.9 & 86.7 & 52.9 & 24.9 \\

\midrule
\rule{0pt}{10pt} $\rm \Ours^{P}$ (R50)  &  54.3 & 80.9 & 93.9 & \textbf{54.5} & \textbf{27.4} & 52.4 & 77.5 & 90.2 & \textbf{55.7} & 28.5 & 50.0 & 74.0 & 86.4 & \textbf{56.9} & \textbf{29.5} \\
\rule{0pt}{10pt} \Ours (R50)  &  54.3 & 76.4&87.6 & 51.4 & 24.0 & 51.8 & 73.3 & 84.1 & 52.4 & 24.8 &  49.3 & 70.1 & 80.6 & 53.3 & 25.7 \\
\rule{0pt}{10pt} \Ours (H48)  & \textbf{49.7} & \textbf{70.2} & \textbf{81.2} & 50.5 & 23.8 & \textbf{47.6} & \textbf{64.3} & \textbf{74.4} & 55.3 & \textbf{28.5} & \textbf{44.9} & \textbf{64.3} & \textbf{74.7} & 55.3 & 28.5     \\
\bottomrule

\end{tabular}

%% file: tables/supp_pdhuman2.tex
\begin{tabular}{lp{57pt}<{\centering}p{47pt}<{\centering}p{47pt}<{\centering}p{47pt}<{\centering}p{47pt}<{\centering}|p{57pt}<{\centering}p{47pt}<{\centering}p{47pt}<{\centering}p{47pt}<{\centering}p{47pt}<{\centering}}
\toprule
\multirow{2}{*}{\textbf{Methods}} & \multicolumn{5}{c}{PDHuman~($\tau=1.8$)} & \multicolumn{5}{c}{PDHuman~($\tau=1.4$)}
 \\ \cmidrule{2-11}
&PA-MPJPE$\downarrow$ & MPJPE$\downarrow$  & PVE$\downarrow$ & mIoU$\uparrow$ &P-mIoU$\uparrow$& PA-MPJPE$\downarrow$ & MPJPE$\downarrow$ & PVE$\downarrow$ & mIoU$\uparrow$ &P-mIoU$\uparrow$ \\ \midrule 

\rule{0pt}{10pt} HMR~(R50)~\cite{hmr} & 53.9 &  79.0 & 92.4 & 53.6 & 25.1 & 49.2 & 73.3 & 85.9 & 57.3 & 28.2\\

\rule{0pt}{10pt} HMR-$f$~(R50)~\cite{hmr}&  53.4 & 78.3 & 91.8 & 50.4 & 24.6  & 48.8 & 72.7 & 85.3 & 54.6 & 28.1 \\

\rule{0pt}{10pt} SPEC~(R50)~\cite{spec}  &  56.8 & 81.8 & 95.1 & 40.1 & 17.1 & 51.8 & 75.4 & 87.9 & 37.4 & 15.3   \\

\rule{0pt}{10pt} CLIFF~(R50)~\cite{cliff} & 56.7 & 83.6 & 97.3 & 56.5 & 29.1 & 51.6 & 76.9 & 89.7 & 60.2 & 32.7\\

\rule{0pt}{10pt} PARE~(H48)~\cite{pare}  & 56.8 & 83.9 & 103.0 & 52.8 & 24.3 & 51.8 & 78.5 & 96.6 & 56.6 & 27.5 \\

\midrule
\rule{0pt}{10pt} GraphCMR~(R50)   &  53.2 & 74.2 & 85.2 & 52.7 & 25.3 & 48.7 & 69.1 & 79.4 & 56.4 & 28.6 \\

\rule{0pt}{10pt} FastMetro(H48)~\cite{fastmetro}   &  49.4 & 71.1 & 81.1 & 55.5 & 27.0 & 45.0 & 65.8 & 75.2 & 59.7 & 31.0  \\

\midrule
\rule{0pt}{10pt} $\rm \Ours^{P}$ (R50)  & 47.1 & 69.8 & 81.6 & \textbf{58.7} & \textbf{30.7} & 43.2 & 65.2 & 76.5 & \textbf{61.3} & \textbf{32.6}  \\
\rule{0pt}{10pt} \Ours (R50)  & 45.9 & 66.0 & 75.9 & 54.8 & 26.8 & 41.9 & 61.5 & 70.9 & 57.2 & 28.2 \\
\rule{0pt}{10pt} \Ours (H48)  & \textbf{42.1} & \textbf{60.7} & \textbf{70.4} & 56.8 & 29.5 & \textbf{39.4} & \textbf{56.6} & \textbf{69.6} & 58.3 & 29.9    \\
\bottomrule
\end{tabular}

%% file: tables/supp_spec_mtp.tex
\begin{tabular}{lccccc|ccccc|ccccc}
\toprule
\multirow{2}{*}{\textbf{Methods}} & \multicolumn{5}{c}{SPEC-MTP~($\tau=1.8$)}& \multicolumn{5}{c}{SPEC-MTP~($\tau=1.4$)}
&\multicolumn{5}{c}{SPEC-MTP~($\tau=1.0$)}\\ \cmidrule{2-16}
&PA-MPJPE$\downarrow$ & MPJPE$\downarrow$  & PVE$\downarrow$ & mIoU$\uparrow$ &P-mIoU$\uparrow$& PA-MPJPE$\downarrow$ & MPJPE$\downarrow$ & PVE$\downarrow$ & mIoU$\uparrow$ &P-mIoU$\uparrow$& PA-MPJPE$\downarrow$ & MPJPE$\downarrow$ & PVE$\downarrow$ &  mIoU$\uparrow$  & P-mIoU$\uparrow$ \\ \midrule 

\rule{0pt}{10pt} HMR~(R50)~\cite{hmr}  & 73.9 & 121.4 & 145.6 & 48.8 & 16.0 & 73.1 & 112.5 & 135.7 & 51.1 & 20.0 & 69.6 & 111.8 & 135.7 & 50.5 & 21.8\\

\rule{0pt}{10pt} HMR-$f$~(R50)~\cite{hmr}& 72.7 & 123.2 & 145.1 & 52.3 & 21.0 & 72.1 & 113.3 & 135.5 & 51.9 & 21.9 & 69.1 & 112.8 & 136.3 & 52.5 & 24.8 \\

\rule{0pt}{10pt} SPEC~(R50)~\cite{spec}  &   76.0 & 125.5 & 144.6 & 49.9 & 18.8 & 72.4 & 114.0 & 134.3 & 49.3 & 19.5 & 67.4 & 110.6& 132.5 & 49.1 & 21.2  \\

\rule{0pt}{10pt} SPEC~*~(R50)~\cite{spec}  &  -& -& -& -& -& -& -& -& -& -& 71.8& 116.1&  136.4 & -& - \\

\rule{0pt}{10pt} CLIFF~(R50)~\cite{cliff} & 74.3 & 115.0 & 132.4 & 53.6 & 23.7 & 70.2 & 107.0 & 126.8 & 52.0 & 22.1 & 67.4 & 108.7 & 130.4 & 51.9 & 23.4\\

\rule{0pt}{10pt} PARE~(H48)~\cite{pare} & 74.2 & 121.6 & 143.6 & 55.8 & 23.2 & 71.6 & 112.7 & 137.2 & 55.1 & 22.4 & 68.5 & 113.5 & 139.6 & 55.3 & 25.1  \\

\midrule
\rule{0pt}{10pt} GraphCMR~(R50)   & 76.1 & 121.1 & 133.1 & 56.3 & 23.4 & 74.4 & 114.9 & 129.5 & 52.6 & 20.8 &  70.2 & 112.7 & 127.8 & 51.7 & 22.0\\

\rule{0pt}{10pt} FastMetro(H48)~\cite{fastmetro}   & 75.0   & 123.1  & 137.0 & 53.5  & 20.5 & 70.8 & 112.3 & 128.0 & 52.4 & 20.6 & 66.3 & 110.2 & 126.5 & 51.8 & 22.6   \\

\midrule
\rule{0pt}{10pt} $\rm \Ours^{P}$ (R50)  & 72.9 & 117.7 & 138.2 & 54.7 & 22.4 & 70.5 & 108.1 & 129.4 & 53.9 & 21.5 &  68.4 & 110.2 & 134.3 & 54.7 & 24.2 \\
\rule{0pt}{10pt} \Ours (R50)  & 74.0 & 122.1 & 135.6 & 58.9 & 24.9 & 70.3 & 111.1 & 126.0 & 56.9 & 22.0 & 66.9 & 109.6 & 124.4 & 56.5 & 23.4 \\
\rule{0pt}{10pt} \Ours (H48)  & \textbf{67.4} & \textbf{114.6} & \textbf{126.7} & \textbf{62.6} & \textbf{30.4}   & \textbf{66.5} & \textbf{106.1} & \textbf{120.1} & \textbf{59.9} & \textbf{26.6} & \textbf{65.8} & \textbf{108.2} & \textbf{121.9} & \textbf{58.5} & \textbf{27.0} \\
\bottomrule
\end{tabular}

%% file: tables/supp_humman.tex
\begin{tabular}{lccccc|ccccc|ccccc}
\toprule
\multirow{2}{*}{\textbf{Methods}} & \multicolumn{5}{c}{HuMMan~($\tau=1.8$)}& \multicolumn{5}{c}{HuMMan~($\tau=1.4$)}
&\multicolumn{5}{c}{HuMMan~($\tau=1.0$)}\\ \cmidrule{2-16}
&PA-MPJPE$\downarrow$ & MPJPE$\downarrow$  & PVE$\downarrow$ & mIoU$\uparrow$ &P-mIoU$\uparrow$& PA-MPJPE$\downarrow$ & MPJPE$\downarrow$ & PVE$\downarrow$ & mIoU$\uparrow$ &P-mIoU$\uparrow$& PA-MPJPE$\downarrow$ & MPJPE$\downarrow$ & PVE$\downarrow$ &  mIoU$\uparrow$  & P-mIoU$\uparrow$ \\ \midrule 

\rule{0pt}{10pt} HMR~(R50)~\cite{hmr} & 30.2 & 43.6 & 52.6 & 65.1 & 39.5 & 31.9 & 45.0 & 39.5 & 66.6 & 39.9 & 30.0 & 44.1 & 50.7 & 66.6 & 39.5 \\

\rule{0pt}{10pt} HMR-$f$~(R50)~\cite{hmr}& 29.9 & 43.6 & 53.4 & 62.7 & 34.9 & 31.3 & 45.0 & 53.3 & 66.6 & 39.9 & 29.8 & 44.1 & 50.7 & 66.6 & 39.5 \\

\rule{0pt}{10pt} SPEC~(R50)~\cite{spec}  & 31.4 & 44.0 & 54.2 & 51.4 & 24.6 & 33.1 & 46.1 & 41.7 & 46.0 & 19.2 &   31.2 & 44.8 & 51.6 & 42.2 & 16.6 \\

\rule{0pt}{10pt} CLIFF~(R50)~\cite{cliff} &  28.6 & 42.4 & 50.2 & 68.9 & 44.7 & 30.3 & 43.3 & 51.2 & 70.2 & 44.9 & 28.3 & 42.3 & 48.5 & 70.6 & 44.5 \\

\rule{0pt}{10pt} PARE~(H48)~\cite{pare} & 32.6 & 53.2 & 65.5 & 64.5 & 38.3  & 33.6 & 53.3 & 66.2 & 65.1 & 38.0 & 32.2 & 53.1 & 64.6 & 65.0 & 37.6\\

\midrule
\rule{0pt}{10pt} GraphCMR~(R50)   & 29.5 & 40.6 & 48.4 & 61.6 & 37.5 & 30.3 & 40.6 & 48.2 & 62.6 & 37.6 & 29.3 & 40.2 & 46.3 & 62.8 & 37.0 \\

\rule{0pt}{10pt} FastMetro(H48)~\cite{fastmetro}   & 26.3 & 38.8 & 45.6 & 68.3 & 45.2 &   27.8 & 39.9 & 46.6 & 69.9 & 45.7 & 26.5 & 38.5 & 43.6 & 70.0 & 45.3 \\

\midrule
\rule{0pt}{10pt} $\rm \Ours^{P}$ (R50)  &  24.4 & 36.7 & 45.9 & 70.4 & \textbf{45.5} & 26.2 & 37.6 & 45.6 & 70.4 & 45.3 & 25.6 & 37.7 & 43.7 & 70.8 & 45.2 \\
\rule{0pt}{10pt} \Ours (R50)  &  25.5 & 36.7 & 43.4 & 67.0 & 38.4 & 25.6 & 36.5 & 42.5 & 70.4 & 42.7 & 24.2 & 35.2 & 40.4 & 70.7 & 42.4 \\
\rule{0pt}{10pt} \Ours (H48)  & \textbf{22.3} & \textbf{32.6} & \textbf{40.0} & \textbf{71.2} & 45.1 & \textbf{24.1} & \textbf{33.8} & \textbf{40.7} & \textbf{72.2} & \textbf{47.9} & \textbf{23.0} & \textbf{33.0} & \textbf{38.7} & \textbf{73.2} & \textbf{47.4}    \\
\bottomrule
\end{tabular}